\documentclass[journal,twocolumn,twoside,letter]{IEEEtran}
\ifCLASSINFOpdf
\else
\fi
\usepackage{ifpdf}
\usepackage{amsmath,epsfig}
\usepackage{hyperref}
\usepackage{epstopdf}
\usepackage{subfigure}
\usepackage[ruled,vlined]{algorithm2e}
\usepackage{multirow}
\usepackage{color,soul}
\usepackage{amssymb}
\usepackage{gensymb}
\usepackage{float}
\usepackage{dcolumn}
\usepackage{url}
\usepackage{cite}
\usepackage{kpfonts}
\usepackage{stackengine}
\usepackage{calc}
\newcolumntype{d}[1]{D{.}{\cdot}{#1} }

\begin{document}
%
\title{Breast Cancer: Model Reconstruction and Image Registration from Segmented Deformed Image using Visual and Force based Analysis  }
%
%
%

\author{Shuvendu~Rana,~\IEEEmembership{Member,~IEEE,}
 		Rory Hampson,
        and~Gordon Dobie
\thanks{S. Rana, R. Hampson and G. Dobie are with the Centre for Ultrasonic Engineering, Department
of EEE, University of Strathclyde, Glasgow, Scotland, UK (e-mail: shuvendu@ieee.org, rory.hampson@strath.ac.uk, gordon.dobie@strath.ac.uk). This work is funded by the EPSRC under grant number EP/Po11276/1 and supported by Pressure Profile Systems Inc.}
\thanks{Copyright (c) 2019 IEEE. Personal use of this material is permitted. However, permission to use this material for any other purposes must be obtained from the IEEE by sending a request to pubs-permissions@ieee.org.}}

%
%

\markboth{IEEE TRANSACTIONS ON MEDICAL IMAGING, VOL. XX, NO. XX, XXX 2019}%
{{Rana  \MakeLowercase{\textit{et al.}}: B\MakeLowercase{reast} C\MakeLowercase{ancer:} M\MakeLowercase{odel} R\MakeLowercase{econstruction and} I\MakeLowercase{mage} R\MakeLowercase{egistration from} S\MakeLowercase{egmented} D\MakeLowercase{eformed} I\MakeLowercase{mage using} V\MakeLowercase{isual and} F\MakeLowercase{orce based} A\MakeLowercase{nalysis  }}}
%



\maketitle

\begin{abstract}
Breast lesion localization using tactile imaging is a new and developing direction in medical science. To achieve the goal, proper image reconstruction and image registration can be a valuable asset. 
In this paper, a new approach of the segmentation-based image surface reconstruction algorithm is used to reconstruct the surface of a breast phantom. In breast tissue, the sub-dermal vein network is used as a distinguishable pattern for reconstruction.  The proposed image capturing device contacts the surface of the phantom, and surface deformation will occur due to applied force at the time of scanning. A novel force based surface rectification system is used to reconstruct a deformed surface image to its original structure. For the construction of the full surface from rectified images, advanced affine scale-invariant feature transform (A-SIFT) is proposed to reduce the affine effect in time when data capturing. Camera position based image stitching approach is applied to construct the final original non-rigid surface. 
  The proposed model is validated in theoretical models and real scenarios, to demonstrate its advantages with respect to competing methods. The result of the proposed method, applied to path reconstruction, ends with a positioning accuracy of 99.7\%.

\end{abstract}


%
\IEEEpeerreviewmaketitle


\section{Introduction}
\label{sec:intro}

\IEEEPARstart{B}{reast} cancer is one of the most common causes of death and public fear in today's clinical environment.
An estimated 1.38 million women worldwide were diagnosed with breast cancer in 2008, accounting for nearly a quarter (23\%) of all cancers diagnosed in women (11\% of the total in men and women). During that year, it was estimated that breast cancer was responsible for approximately 460,000 deaths worldwide~\cite{globalcancer}.
In UK breast cancer is typically detected through self-examination which induces a visit to the General Practitioner, or through the screening of women over the age of fifty using mammography where only about 8\% of patients referred to secondary care centres have cancer~\cite{clinic}. This proposal considers a new technique to support screening based on tactile imaging. Tactile imaging in a primary care setting has the potential to significantly improve the accuracy of these referrals, reducing patient anxiety through efficient diagnosis while reducing the financial strain caused by unnecessary referrals to secondary care.

\begin{figure}[h!t!]
\centering
\centering

\subfigure[SureTouch Device]{\includegraphics[width=.48\linewidth]{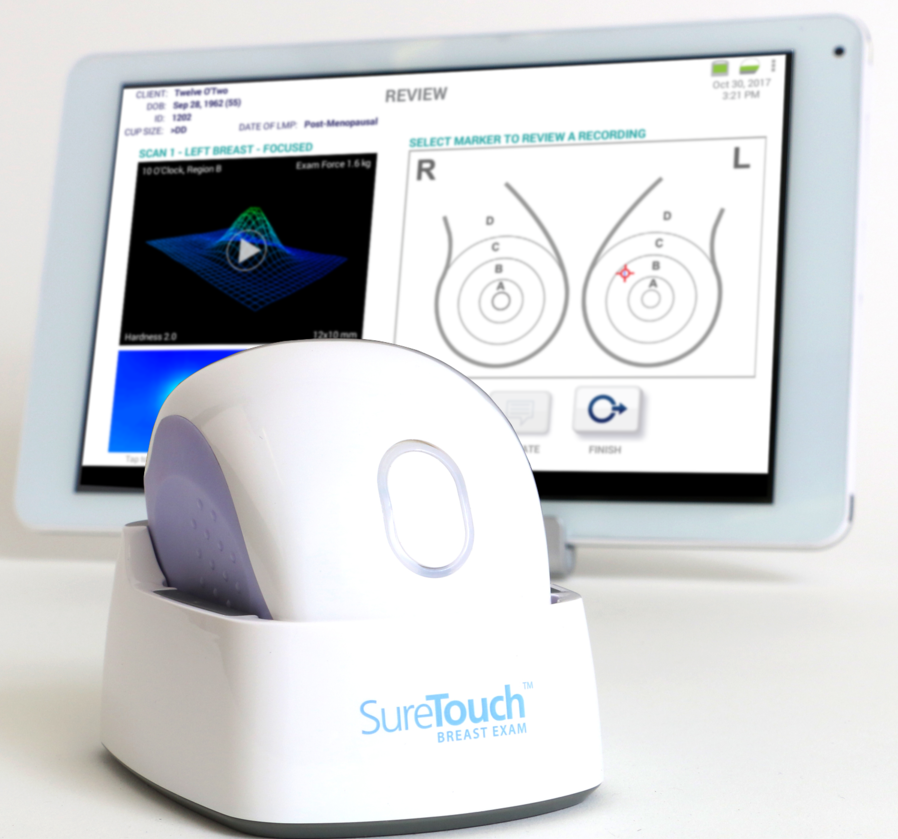}  \label{sta}}
\subfigure[Hardness (lump) detection in breast]{\includegraphics[width=.575\linewidth]{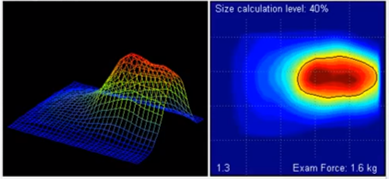}  \label{stb}}
\subfigure[Localization of the lump (manually)]{\includegraphics[width=.375\linewidth]{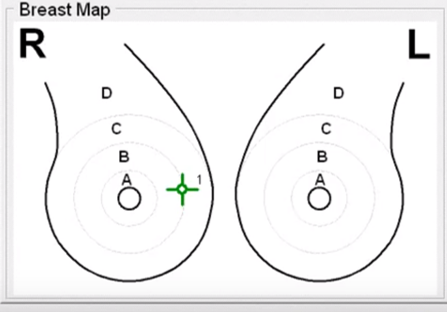}  \label{stc}}
\caption{Sample images of SureTouch device and output~\cite{mechanical,bcancer}}
\label{fig:st}
\end{figure}

Unlike mammography, which provides a complete image of a breast~\cite{mam1,mam2,mam3} using radiation, tactile imaging sensors are scanned over the breast in an noninvasive manner; producing a real time feed of the pressure profile under the sensor~\cite{mechanical,bcancer} as shown in  Fig.~\ref{fig:st}.
 This temporal image feed is more difficult to interpret than a global image of the breast as the data has no spatial context reference. In this scenario, optical sensors, such as embedded cameras can be used to localise an image on to the human body. 

In this scenario, breast tissue consists of numerous veins to feed the mammary glands, that can be imaged with near field infra-red (NFIR) with wavelengths in the range 650nm to 930nm~\cite{nir,nir2}.  That vascular network presents an interesting opportunity for absolute localisation of images with respect to the breast.
  The accurate positioning of the tactile sensor relative to the breast would enable spatial mapping of the tactile data, facilitating the production of a global stress image of the breast. From a clinical perspective, this global image would be a far more effective means of representing data, simplifying interpretation and enhancing diagnosis.

In this research an Infra-red (IR) camera is used as basis of the prototype model to scan the breast phantom. As the images are captured in time with capturing the tactile data, a full model can be constructed for localization of the current position using non-rigid reconstruction.  Also, in time of scanning application of force creates deformation in the breast image and the deformation is created by the camera scanning surface. Thus, the non-rigid reconstruction needs a deformation reconstruction and camera localization method.

In recent literature of non-rigid reconstruction, Agudo et al. proposed the SLAM (Simultaneous Localization And Mapping) for elastic surface~\cite{agudo1}, where they have used the fixed position of the object boundary for image registration. They then later proposed a free boundary condition approach~\cite{agudo3,agudo2} as an extension of their previous work.  In both cases they have used a FEM (finite element model)~\cite{fem1} approach to estimate the deformation with a partially fixed rigid camera. They have used the Navier's equation over FEM to estimate the deformation force. During of deformation, the surface strain and the material stresses are estimated using an EKF (extended Kalman filter)~\cite{ekf1,ekf2}. This physics-based approach requires an initialization step (say beginning static model). In this model, a noisy environment may create an accuracy error, besides the fact that errors may accumulate over the considered time frame. As such, these approaches~\cite{agudo1,agudo2,agudo3}  may not be able to exploit the mechanical constraints to cover a large deformation range. Additionally, rigid reconstruction techniques fail when applied directly to time-deforming objects. 
Shape from Template (SFT) is another method to reconstruct deformation in real linear material~\cite{sft1} and non linear material~\cite{sft2}. Haouchine et al. proposed a SFT method for capturing a 3D elastic surface~\cite{sft1} and create augmented 3D objects using a single viewpoint as a reference view. The authors quantify the material elasticity by using non-linear solvers based on the Saint Venant-Kirchhoff model. A better approach is presented by Malti and Herzet using Sparse Linear Elastic-SFT instead of classic SFT. The authors used a relative ground truth of the original 3D object to match with the observed relative deformation by calculating the spatial domain of non-zero deforming forces.
Usually, the FEM covers the structured deformation having a sequence~\cite{JMIV17}. But in our proposed scheme, the deformation does not have any sequence because application of force in time of scanning is responsible for the stretching deformation. Also, Application of human force may not be modelled in any sequence pattern of structure. As a result, the FEM technique may not provide solution of this unstructured large deformation.
%
In order to overcome this limitation, Non-rigid structure from motion (NRSfM)~\cite{nrsfm1,nrsfm2} techniques have been proposed to recover the 3D shape of non-rigid objects over time. In this scenario, Garg et al. proposed the structure from motion (SfM) technique, which exploits motion cues in a batch of frames~\cite{nrsfm1} and, have been applied to dense non-rigid surface reconstruction (). In another research work, Sepehrinour and Kasaei have used the NRSfM and optical flow method for reconstruction of 3D objects from video sequences~\cite{nrsfm2}. Generally, perspective projection is considered to be a more realistic model and is ideally suited to a wide range of cameras. Currently, NRSfM technique have focused on an orthographic projection camera model, due to its simplicity but, perspective projection yields equations that are complex and often non-linear. Therefore to simplify the calculations, some approximations are applied on the perspective projection model, such that it can be reduced to an orthographic projection. Orthographic reconstruction of non-rigid surfaces has been done by a singular value decomposition algorithm and using the orthogonal characteristics of the rotation matrix~\cite{nrsfm1,nrsfm2}, but true perspective reconstruction of non-rigid surfaces, due to the high complexity and the large number of unknowns, seemed impossible. 
Later, Yu et al.~\cite{dense1} proposed a template based non-rigid 3D reconstruction from a stationary object deformation. The authors have used the dense direct matching template based direct approach to deformable shape reconstruction. In this scenario a template-based method and a feature track based method are used to generate the template from the monocular camera views.  Using the concept of the 2D RGB image, Innmann et al. proposed a 3D volumetric approach to map the observed deformation into a 3D model~\cite{dense3}. The main procedure is to match the scale invariant feature transform (SIFT) key points with the original 3D model to estimate the current position of the object, but the authors have used a constructed 3D model to map with the current deformation. In this approach, they~\cite{dense3} have used a rigid movement of a camera for 3D construction, additionally a fixed distance object location system is used on the deforming object to find the appropriate positioning.
 Later, Agudo et al. proposed a real-time 3D reconstruction technique for nonrigid shapes in~\cite{CUIV16}. Here the reconstruction is made in the time of scanning with a distance camera. But, in this approach~\cite{CUIV16} the reconstruction of the deformation due to the camera scanning plane may not work. Also, the stretching deformation is not static due to the handheld camera. As a result, this approach may not solve the proposed problem.
Very recently, Newcombe et al. proposed a dens reconstruction method using the hand held  RGB-D cameras for non-rigid objects~\cite{dense2}.  Their main proposal was to correct the point to plane mapping error in the observed depth map. The authors have used the sparse feature-based method which is fused into a canonical space using an estimated volumetric warp field, which removes the scene motion, and a truncated signed distance function volume reconstruction is obtained. They have used a fixed platform with a hand-held camera to identify the volume or spatial position of the object. Since they omitted the RGB stream which contains the global features, their method fails to track surfaces in specific types of topological changes and it is also prone to drift. In this approach, depth is the main part for 3D model construction. Hence, application of this method~\cite{dense2} may not fulfil the requirement of the proposed method. 

The existing literature summarizes that the majority of the existing methods work with either a rigid object or a rigid camera and a few literatures defines the rigid camera and rigid object from a sudden distance.   Moreover, for non-rigid camera motion and a non-rigid object, a pre-advised structure of the object is required for registration. Hence, both non-rigid registration in time of the captured non-rigid object is the most challenging task. In this work, the reference segmented non-rigid deformed images are used to construct the surface. Hence, reconstruction is one of the major concerns for model construction. To overcome the issues, force-based reconstruction is carried out to rectify the structure of the surface. A visual reconstruction is carried out to estimate the camera position to reconstruct the deform structure to its reference. To obtain the projection based matching in time of the visual reconstruction Affine SIFT is modified according to the requirement and used to achieve the best performance. 

In summary, a deformation model feature estimation is formulated by proposing a modified Affine-SIFT model to estimate the camera position using SfM for deformed and original coefficients in Sec.~\ref{s2}. Visual and force based reconstruction and model reconstruction are explained in Sec.~\ref{s3} with an experiment set-up in Sec.~\ref{s4}. The accuracy of the proposed method is presented in Sec.~\ref{s5} and finally the paper is concluded in Sec.~\ref{s6}.


\section{Deformation based Feature Estimation}\label{s2}
\begin{figure}[h!t!]
\centering
\centering
\includegraphics[width=.9\linewidth]{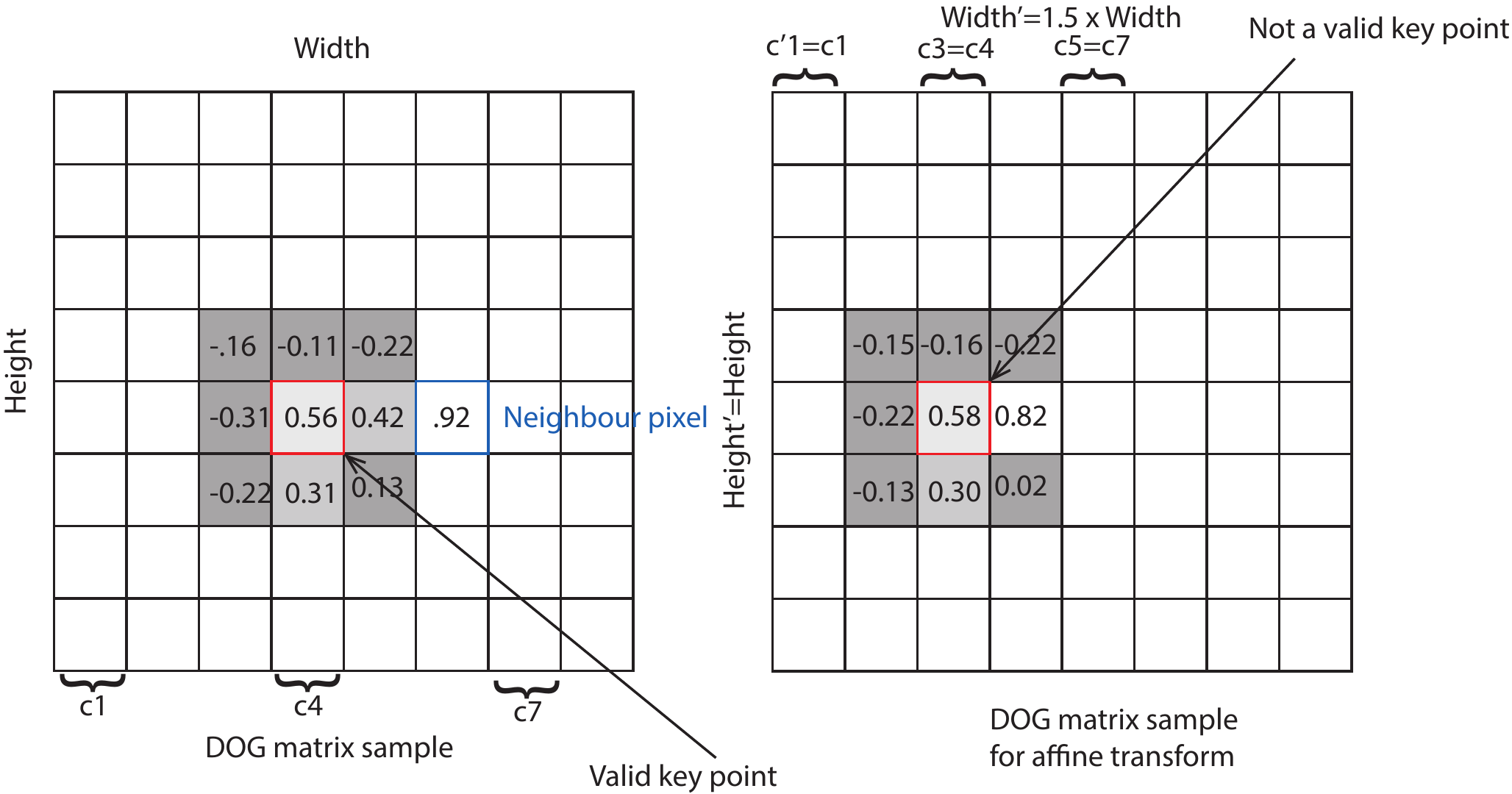}

\caption{Effect of Affine transformation in DOG matrix and the key point selection for SIFT. }
\label{fig19}
\end{figure}

Model construction and image stitching is carried out by analysing the overlapping regions of the multi-view images, similar to those in Fig.~\ref{fig20}. To estimate that, spatial feature matching is the common method~\cite{f_match}. In this scenario, the scale invariant feature transformation (SIFT)~\cite{sift1} is one of the most efficient feature estimations for identifying similar regions~\cite{sift_best}. SIFT features are estimated by analysing the difference of Gaussian (DOG) of the Gaussian pyramid of a selected octave~\cite{sift1}, and scaling does not affect the characteristics of the Gaussian pyramid~\cite{sift1,asift}. It is observed that the change in aspect ratio changes the representation of the DOG matrix and create dissimilarity~\cite{asift}  in the feature point location compare to the original ones as shown in Fig.~\ref{fig19}.
In this work, a hand-held camera is used to take the surface vein photograph of the breast phantom. At the time of image acquisition, the camera will touch the surface of the breast phantom resulting in deformation of the vein pattern due to the application of force required for tactile imaging (as the hardness of the breast phantom is close to the real breast). For the re-construction of the breast phantom surface, the accurate position of each sample images needs to be identified using suitable feature extraction and similar regions identification.  In this scenario, a suitable feature extraction method is presented by analysing the deformations which are described in the subsequent subsections.

\subsection{Deformation Model}\label{def_mod}
\begin{figure}[h!t!]
\centering
\centering
\includegraphics[width=.85\linewidth]{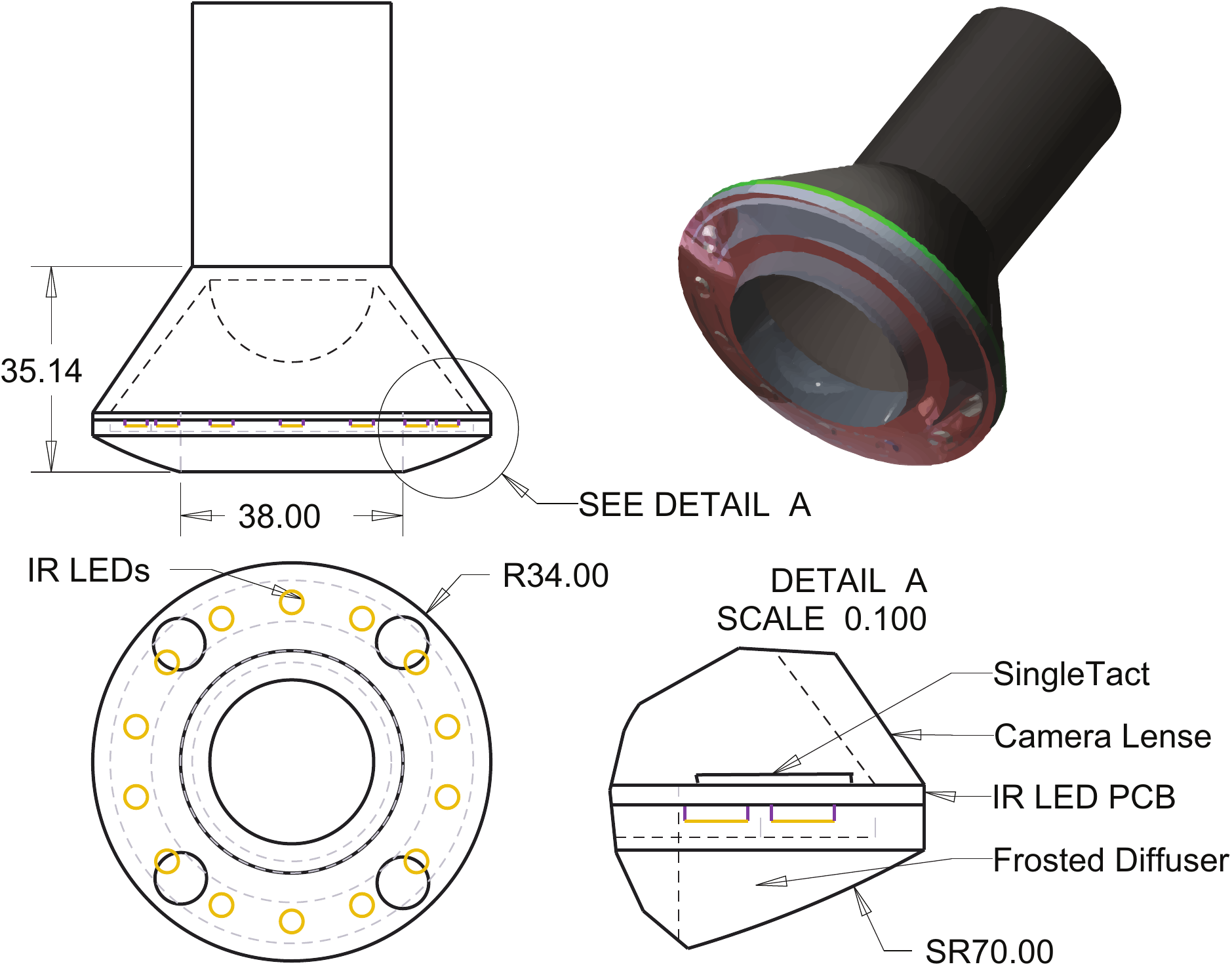}
\caption{Scanner camera and front surface geometry, with assembly shown and IR LED position and SingleTact sensors indicated. (4 single tact sensors are used to measure the force and the application angle. DETAIL A shows the vertical intersection diagram of lens, PCB, 850 nm IR and diffuser.)}
\label{fig4}
\end{figure}

As the camera surface touches the phantom surface the deformation will be formed on the camera view plane. As a flat surface is used for camera view plane as shown in Fig.~\ref{fig4} 
the applied force, measured using four 10N rated SingleTact CS8-10N pressure transducers (PPS Inc, US-CA), during image acquisition will cause a stretching of the structure in the lateral directions. 
\begin{figure}[h!t!]
\centering
\centering
\subfigure[Scaling deformation (where force F1$<$F$<$F2)]{\includegraphics[width=.76\linewidth]{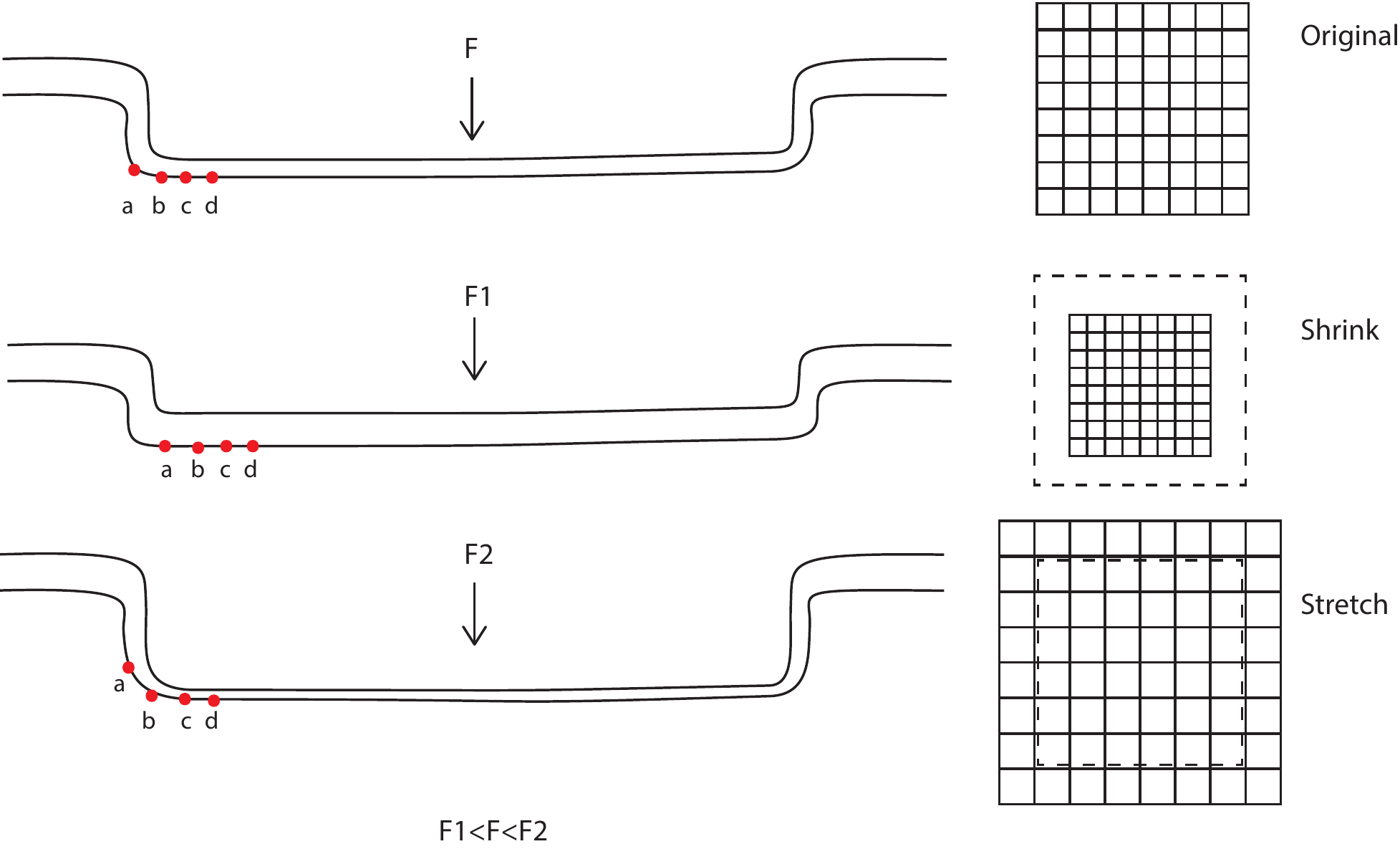} \label{fig5a}}
\subfigure[Affine deformation (where force F1=F2 and F'1$<$F'2)]{\includegraphics[width=.76\linewidth]{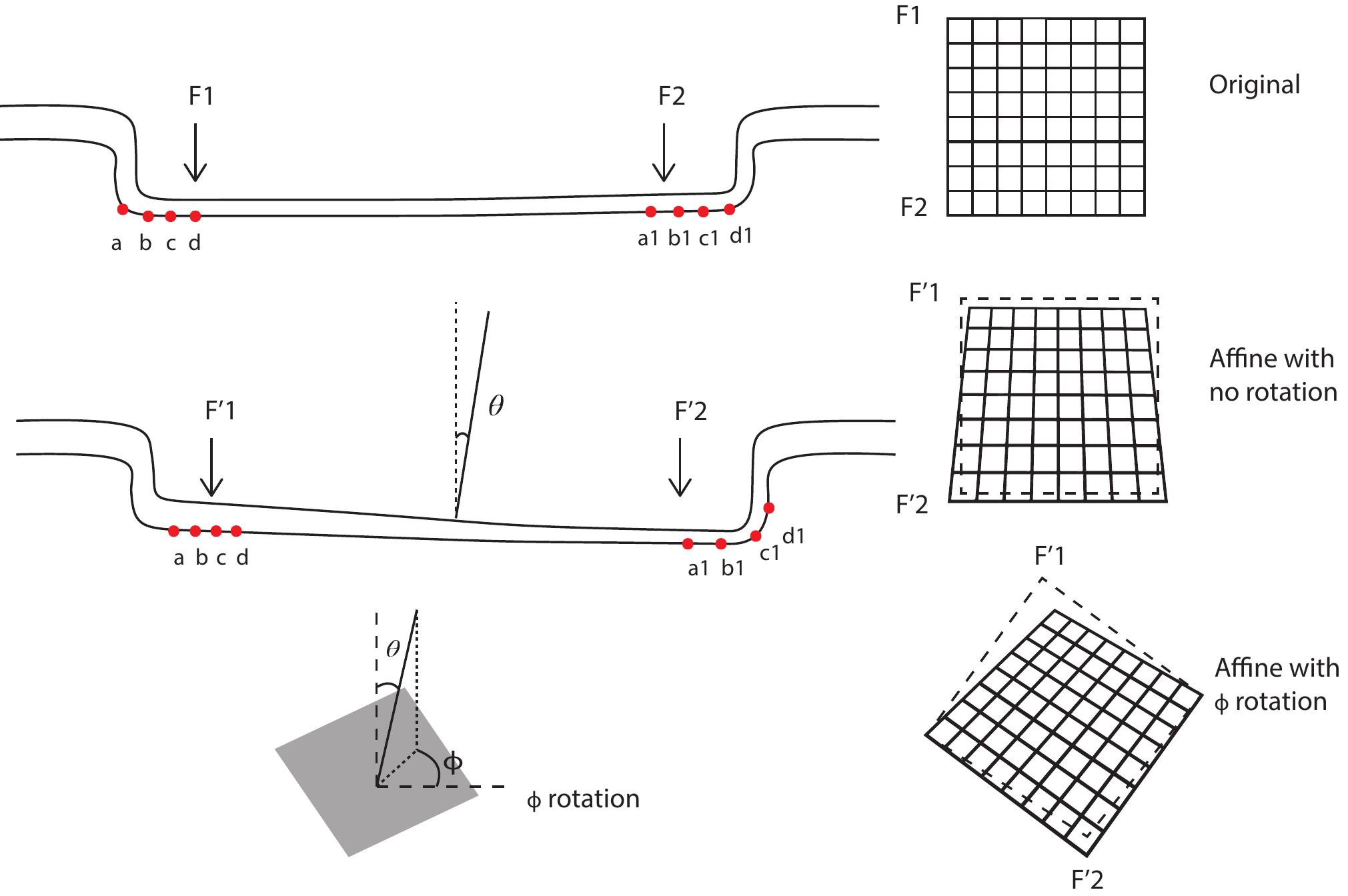} \label{fig5b}}
\caption{Deformation of images due to the application of force. (a) Shows the surface structure due to an application of a normal force with surface markers  `a', `b', `c'. (b) Shows the  surface structure due to application of angular force at angle `$\theta$' with surface markers `a', `b', `c', `a1', `b1', `c1'. The third image shows the structure of the surface with `$\phi$' rotation. }
\label{fig5}
\end{figure}
As the surface is flat, vertical force (with respect to the surface) will create a uniform force distribution at each point in the view plane and cause symmetrical stretching by maintaining the aspect ratio and structure as shown in Fig.~\ref{fig5a}.  An angular offset force will cause different stretching effects at different points as shown in Fig.~\ref{fig5b}.

Here, the stretching ($\mathcal{S}$) and the force ($\digamma$) can be related by (\ref{e5}),
\begin{equation}
\frac{\mathcal{X'}-\mathcal{S}}{\mathcal{X'}}=\upsilon*\digamma
\label{e5}
\end{equation}
where $\mathcal{X'}$ is the structure after applying force and $\upsilon$ is the stretching constant. The stretching constant can be calculated by measuring the \textit{Young's modulus} and other material properties. A detailed stretching and force relation is explained in the next sections.
The calibration of the Young's Modulus (stretching constant) must be done for each new sample, or series of measurements, and can be performed using the proposed device (as described in Section~\ref{pmc}).

In this experiment, the applied force is not always normal to the surface and results in the affine transformation. As discussed earlier, the affine transformation makes significant changes in the Gaussian pyramid. As a result, the extracted SIFT features from the affine deformation do not match with the original image. So an affine feature estimation is necessary to find accurate matching in similar parts of the images.
 
\subsection{Affine Feature Extraction}
An efficient feature estimation technique is required for affine transformation. Affine SIFT is one well known affine feature estimation technique~\cite{asift}.  Here, the tilt parameter is defined by the scaling factor  as in (\ref{e6}),
\begin{equation}
t=\frac{1}{cos(\theta)}
\label{e6}
\end{equation}
where $\theta$ is the tilt angle for the scale factor \cite{asift,asift_direction}.

In this work, the stretching and shrinking can be projected as tilt in camera viewpoint. By experimental observation, the image stretching ratio ($\mathcal{R}$) is $1.13:1$. In the real scenario, the camera can be tilt by $\approx28\degree$ to get the tilt ratio. So, using of standard A-Sift will miss feature points for tilts  $\leq 28\degree$.


\subsubsection{Modified A-SIFT}
  For this work the features need to be extracted from such a latitude angle, that it can cover the maximum ($28\degree$) and minimum tilt ($0\degree$). Thus $\Delta t$ will be the geometric factor of the scale ratio as shown in (\ref{e7}),
 \begin{equation} 
   \Delta t=\sqrt{\mathcal{R}}
   \label{e7}
\end{equation}
where the latitude tilt angle $\theta= cos^{-1}\frac{1}{\Delta t}\approx 20\degree$ to cover the SIFT features for affine transformation. For the longitude or the rotation effect, $\Delta\phi$ is taken as $20\degree$~to form a equilateral triangle in the 3D space for scaling measurement. In this scenario, the SIFT calculation  will be carried out for \{$t=1$,$t=\Delta t | \phi =(0, 20, ...,160)$\}. The total SIFT calculation area will be $(1+cos(20)*9\approx9.46)$ times of original SIFT. The A-SIFT matching overload will be $\approx (9.46)$ if matched with normal SIFT features.

\section{Reconstruction Model}\label{s3}
\begin{figure}[h!t!]
\centering
\centering
\includegraphics[width=.75\linewidth]{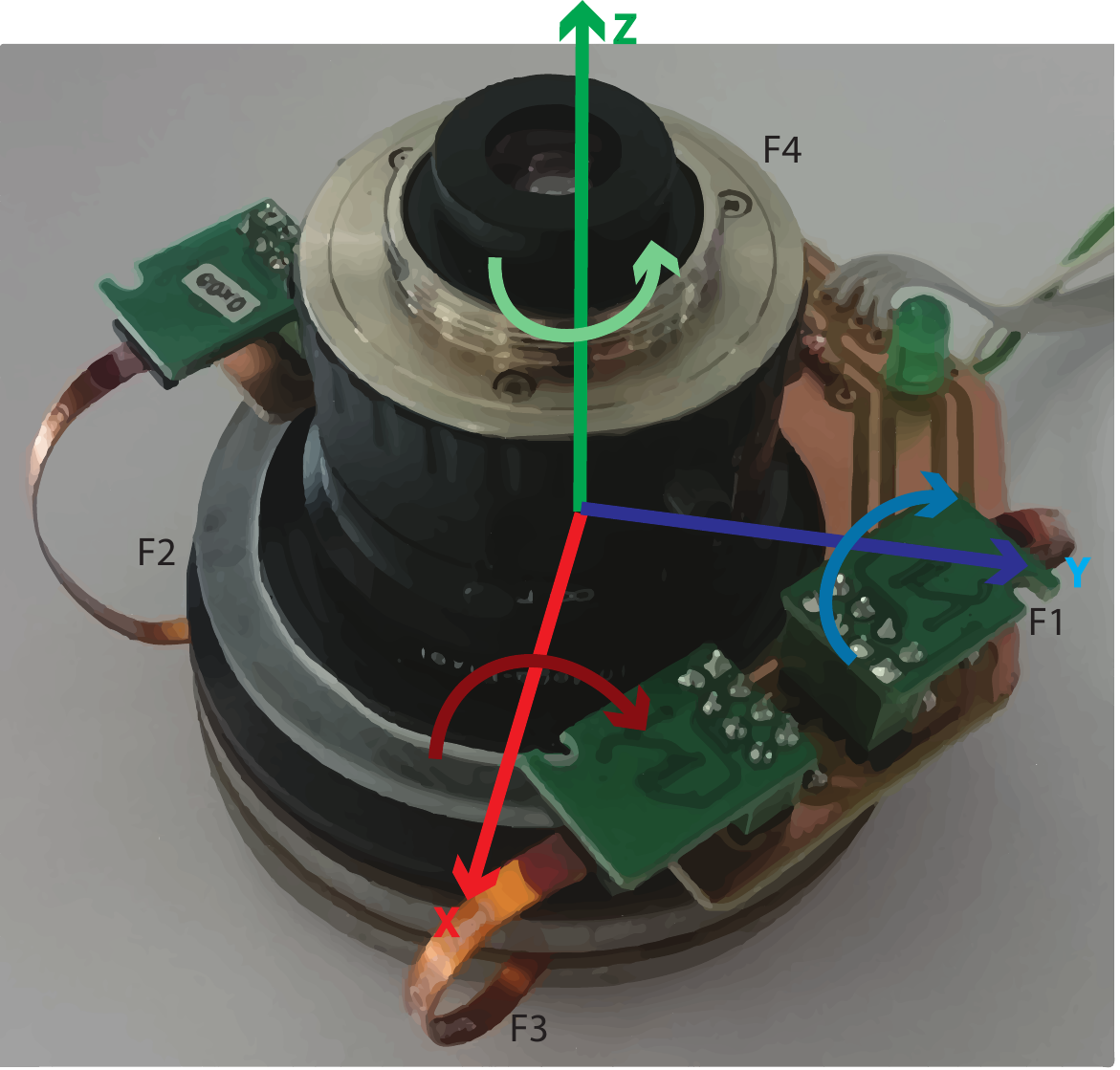}
\caption{Model probe with the four force sensors creating a right handed orthogonal coordinate frame. $\vec{F_4F_3}=X$ axis, $\vec{F_2F_1}=Y$ axis }
\label{fig7}
\end{figure}

Deformation rectification is the most essential part of image reconstruction. As the deformations are caused by the application of force at the time of scanning, structure from motion (SfM)~\cite{sfm1,sfm_asift} based visual models are not sufficient to reconstruct the surface. Thus the deformation needs to be rectified by measuring the applied force and undoing the deformation before applying the visual model.
 In this work, pressure sensors are used to estimate the applied force, and the angle of the probe as shown in the prototype model in Fig.~\ref{fig7}.
  Using this force measurement technique and the implied lateral strains of the phantom, original structure can be reconstructed. 

\subsection{Force Based Deformation Model}\label{fmrm}
As discussed earlier, the application of force causes a stretching effect on the non-rigid surface based on the material \textit{Young’s modulus} as shown in (\ref{e9}),
\begin{equation}
\varepsilon=-\frac{\upsilon}{{E}}\sigma_Z
\label{e9}
\end{equation}
where $\varepsilon$ is the orthogonal lateral strains in both X and Y dimensions, $\upsilon$  is the material Poisson ratio, $E$ is the material \textit{Young’s modulus} and $\sigma_Z$ is the applied axial stress in the Z direction.

From an imaging standpoint, for an axial load with the camera axis coincident with the loading axis, an image feature will stretch proportionally to that load, which is consistent with (\ref{e9}). Additionally, each feature unit will deform by a factor of the applied load as should be expected from (\ref{e9}) and, the change in a feature's radial location, with respect to the camera axis, required to restore the feature given an axial load will also be proportional to the feature's respective stretched radial distance from the camera centre as described in (\ref{e10}).
\begin{equation}
\frac{\Delta r'_i}{r'_i}=-\frac{\upsilon}{{E}}\sigma_Z
\label{e10}
\end{equation}
Where $r'_i$ is the is the radial distance of the $i^{th}$ image feature after applying the axial load and $\Delta r'_i$ is the distortion rectification required to get the undistorted structure. $r_i$ is the undistorted radial distance of $i^{th}$ image feature. So $\Delta r'_i$ can be represented as (\ref{e111}):
\begin{equation}
\Delta r'_i=r_i-r'_i
 \label{e111}
\end{equation}
Using (\ref{e10}) and (\ref{e111}), the relation between $R_i$ and $R'_i$ can be represented as shown in (\ref{e11}).
\begin{equation}
r_i\left | \begin{matrix} 
=r'_i-r'_i 
\begin{array}{@{}c@{}}
{\upsilon}\\\hline{{E}}
\end{array}
\sigma_Z \\
= r'_i(1- \begin{array}{@{}c@{}}
{\upsilon}\\\hline{{E}}
\end{array}\sigma_Z)
\end{matrix} \right.
 \label{e11}
\end{equation}

The stretching relation in (\ref{e11}), can be applied to the wider range of situations where the loading axis is not normal to the breast surface, i.e. the camera is tilted with respect to the structure surface, by realizing that the applied load will not be uniform across the image plane as it would be in the case coincident axes.

  $\digamma_1,\;\digamma_2$ are the forces along the Y axis and $\digamma_3,\;\digamma_4$ are the forces along the X axis. The total force in the Z direction, applied at the centre of the image, can be calculated using (\ref{e11a}).
\begin{equation}
\digamma_Z={\digamma_1+\digamma_2+\digamma_3+\digamma_4}
\label{e11a}
\end{equation}

Additionally the tilt angles of $\theta_x$ \& $\theta_y$ about the X and Y axis respectively can be calculated using (\ref{e11b}).
\begin{equation}
\left. \begin{array}{l} 
\theta_x=sin^{-1}\left(\begin{array}{@{}c@{}} 
{\digamma_2-\digamma_1}\\\hline
{2\kappa S_X}\end{array}
\right)\\
\theta_y=sin^{-1}\left(\begin{array}{@{}c@{}} 
{\digamma_4-\digamma_3}\\\hline {2\kappa S_Y}\end{array}
\right)
\end{array} \right.
\label{e11b}
\end{equation}
Where $S_x$ ans $S_Y$ are the separation between the sensors along the X and Y axis, for this experiment $S_X=S_Y=53mm$.
For a tilt of $\theta$, a load variation across the image plane will be caused by a deviation in the scanner depth into the structure dictated by the material spring constant, $\kappa$, as described in (\ref{e12}).
\begin{equation}
\digamma_Z^{(x,y)}=\digamma_Z+\kappa \mathcal{Z}^{x,y}
\label{e12}
\end{equation}
where $\mathcal{Z}^{x,y}$ is the depth deviation in the Z direction at location (x,y) of the image plane. The load distribution across an image is calculated using the average load at the centre of the image with a position dependent offset related to the tilt angles of the scanner by firstly defining a flat image plane normal vector, $u=[0\;0\;1]^T$. Then, for tilt angle of $\theta$ the tilt normal vector $u'$ can be calculated using (\ref{e13}).
\begin{multline}
u'=\left[ {\begin{array}{*{20}c}
   1 & 0 & 0  \\
   0 & cos(\theta_x) & -sin(\theta_x)  \\
   0 & sin(\theta_x) & cos(\theta_x)  \\
\end{array}} \right]\\
\left[ {\begin{array}{*{20}c}
   cos(\theta_y) & 0 & sin(\theta_y)  \\
   0 & 1 & 0  \\
   -sin(\theta_y) & 0 & cos(\theta_y)  \\
\end{array}} \right]
[u]
\label{e13}
\end{multline}

The depth deviation $\mathcal{Z}^{x,y}$ can  be calculated using (\ref{e14}). 
\begin{equation}
\mathcal{Z}^{x,y}=\frac{-1}{u_z'}(u_x'+u_y'+0)
\label{e14}
\end{equation}

Combining the lateral strain equation (\ref{e11}) and force distribution equation (\ref{e12}), the relation between the force and change of radial pixel location can be represented in a general form for non normal axial loads, based on tilt angle and average load, as shown in (\ref{e15})
\begin{equation}
r^{(x,y)}=r'^{(x,y)}(1- \frac{\upsilon}{{E}\mathcal{A}}\digamma_Z^{(x,y)})
\label{e15}
\end{equation}
where $\mathcal{A}$ is the scanner area. In this experiment the value of $\mathcal{A}=0.0048m^2$. 
Using this analytical equation, the surface structure can be estimated as the equation undoes the warping in the image. However the position and orientation of the surface can not be estimated using the force analysis as no yaw term can be measured. A visual position estimation of the surface will result the complete model construction in this situation.
\subsection{Visual Reconstruction Model}\label{vrm}
For surface construction and image stitching, proper reconstruction of the deformed surface is an important task. In this scenario, camera position estimation and re-projection can be the best possible solution to understand the original surface macro-structure. Hence, camera position can be used to estimate the original structure of the surface. For this, SfM will provide the relative position and orientation of the surface. As the structure of the surface is affine due to the force deformation effect as discussed earlier, modified feature estimation will be used to get matching features for the deformed model.

Using the modified A-SIFT and SfM~\cite{sfm} pipeline, relative camera position can be estimated by defining the rotation and translation vector $[R|T]$.
As the scanning is done by touching and compressing the surface, we can assume that the change of position in the Z direction will define the change of force as in (\ref{e9}). A relative position change estimation can be used to correct the translation matrix for re-projection. To compute the reconstructed structure, the angle $\theta$ needs to be zero, a condition awarded by the deformation model. 

Let $[R_i|T_i]$ define the relative rotation and translation matrix of $i^{th}$ image to project $I_{X,Y,Z}$ positions using the relation as shown in (\ref{e16}) 
\begin{equation}
z\left[ {\begin{array}{*{20}c}
   {x_i }, & {y_i }, & {1 }  \\
\end{array}} \right]
 = 
\left[ {\begin{array}{*{20}c}
   X ,& Y ,& Z  \\
\end{array}} \right]
[R_i |T_i ]\mathcal{C}
\label{e16}
\end{equation}
where (x,y,1) defines the homogeneous location $i^{th}$ image for the camera intrinsic value $\mathcal{C}$. To identify the camera position, re-projection of the points from the corrected non-deformed position will show the original structure.

\begin{figure}[h!t!]
\centering
\centering
\includegraphics[width=.56\linewidth]{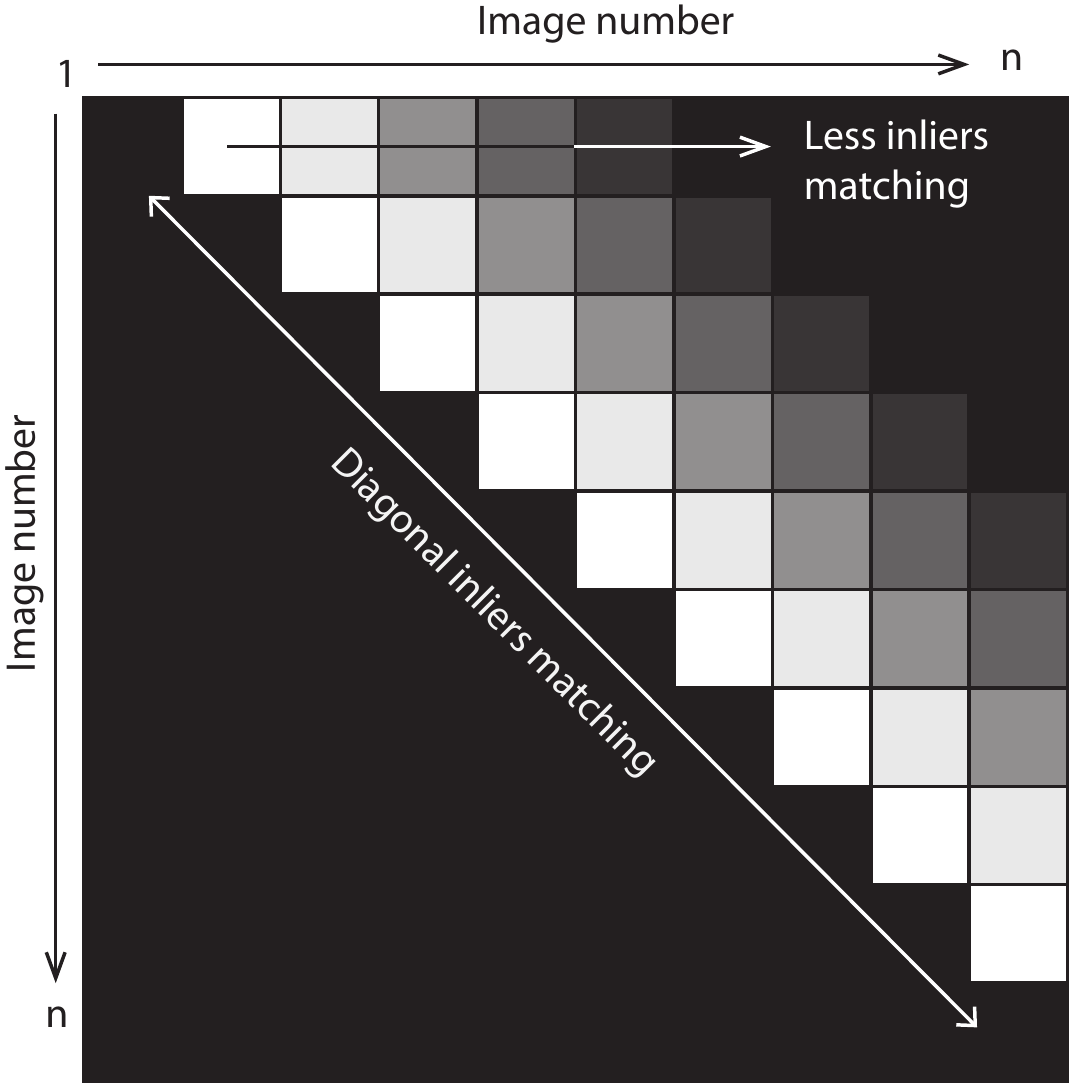}

\caption{Inliers matching for sequential Images. White areas indicate better matching due to physical closeness of images. }
\label{fig18}
\end{figure}

$I_i$ and $I_{i+1}$ are the two images where the image $I_{i+1}$ is deformed.
To get best features, modified A-SIFT is applied on the $i+1^{th}$ image so that better matching is possible with $i^{th}$ image. Using the SfM pipeline~\cite{sfm}, feature matching, and Random sample consensus (RANSAC), we estimate inlier matching from a set of SIFT matching data between $i+1^{th}$ image and $i^{th}$ image that contains outliers~\cite{ransac,ransac1}. Fig ~\ref{fig18} shows a sample of the matching sequences where the white region represents more inliers matching and dark represents the fewer inliers matching.
It can be assumed that for sequential scanning, the majority of matching will occur in the next image and a sufficient number of matching points can be obtained. 
 The eight-point algorithm will provide the fundamental matrix $\mathcal{F}$ using the  inlier points as shown the relation in (\ref{e17}) 
\begin{equation}
[x_{i+1},y_{i+1},1]^T\mathcal{F}_{i+1}[x_{i},y_{i},1]=0
\label{e17}
\end{equation}
where $[x,y,1]$ defines the homogeneous location of matching inliers. Using the camera parameter $\mathcal{K}$ and the fundamental matrix $\mathcal{F}$, relative camera position and orientation of the  $I_{i+1}$ image can be calculated using Single Value Decomposition (SVD) and solution estimation method~\cite{ransac, sfm,epipolar }. For this, $[R_{i+1}|T_{i+1}]$ defines the relative rotation and translation of the camera `i+1'. 

\begin{figure}[h!t!]
\centering
\centering
\includegraphics[width=.7\linewidth]{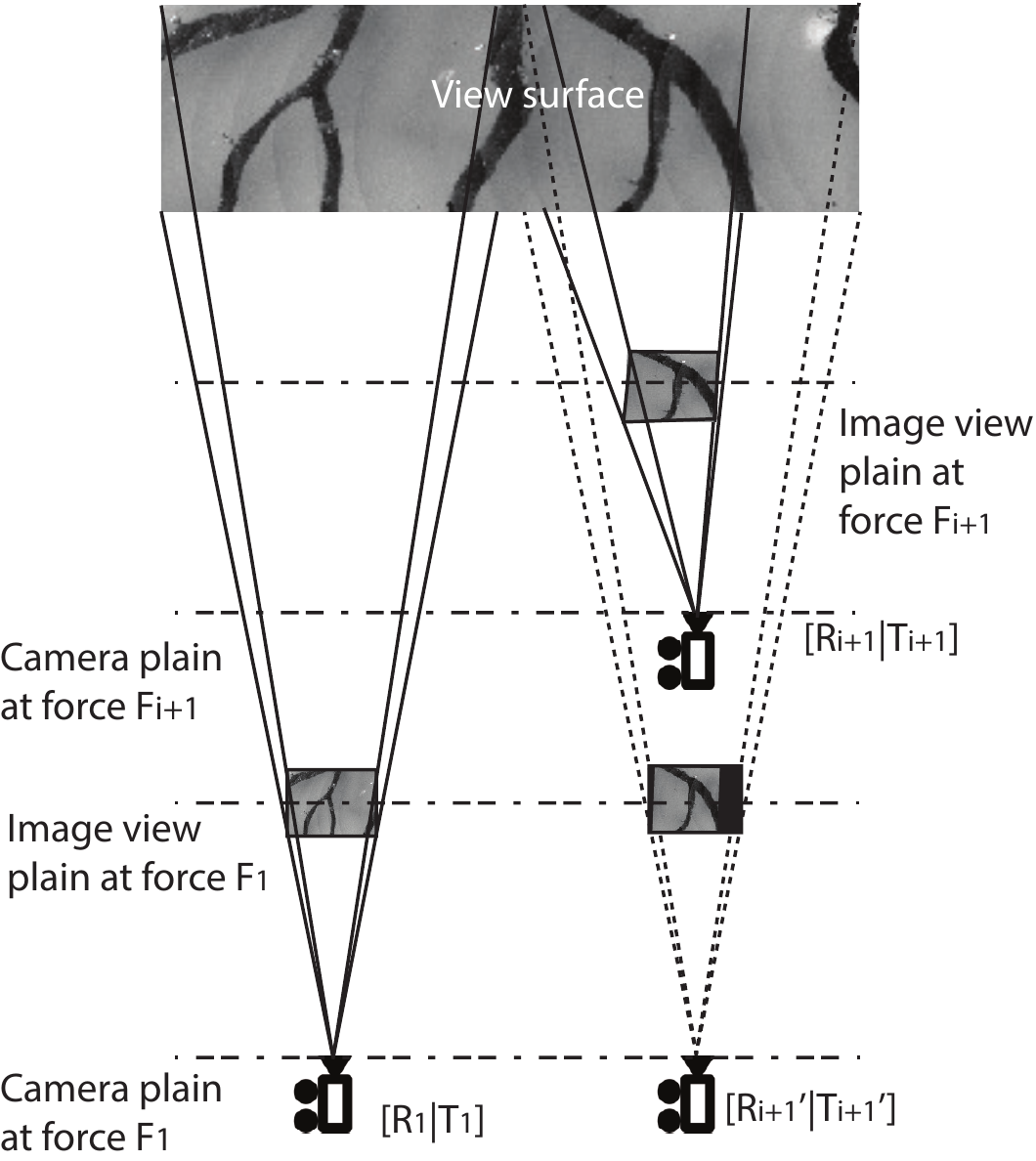}

\caption{Camera plane and view plane re-projection using re-estimation of the rotation and the translation matrix to reduce the force based visual deformation (where force $F_1<F_{i+1}$ and camera plain and view plain are calculated using image deformation analysis).}
\label{fig113}
\end{figure}

To correct the visual deformation correctly the camera position, rotation, and translation should be the same plane of the $i^{th}$ image as shown in Fig.~\ref{fig113}.  

Hence, the corrected R and T matrices can be represented as shown in (\ref{e18})
 \begin{equation}
\begin{array}{l}
 R'\left| {\begin{array}{*{20}c}
    =& R_X (\theta _X )R_Y (\theta _Y )R_Z (\phi )  \\
    =& \left[ {\begin{array}{*{20}c}
   1 & 0 & 0  \\
   0 & 1 & 0  \\
   0 & 0 & 1  \\
\end{array}} \right]\quad\quad\quad  \\
\end{array}} \right.for\;\theta _X  = \theta _Y  = \phi  = 0 \\ 
 T' = [\begin{array}{*{20}c}
   {t_x }, & {t_y }, & 0  \\
\end{array}]\quad\quad\quad\quad\quad\quad\; for\;t_z  = 0 \\ 
 \end{array}
\label{e18}
\end{equation}
where $R'$ and $T'$ are the corrected rotation and the translation respectively. In this case, rotation about the Z axis ($\phi$) is removed to correct yaw rotation effect and the rotations about the X and the Y axis ($\theta_X$ and $\theta_Y$) are removed to correct the affine deformation. The translation vector, ${t_x }$ and ${t_y }$ defines the X and Y translation. Here ${t_z }$ is set to `0' to remove the scaling effect. The image is re-projected using the corrected $[R'|T']$ as described in (\ref{e19})

\begin{equation}
z'[\begin{array}{*{20}c}
   {x'}, & {y'}, & 1  \\
\end{array}]\left|
 {\begin{array}{*{20}c}
 =[&X,&Y,&Z&][R' |T' ]\mathcal{C}\\
  =[&x,&y,&1&]\begin{array}{@{}c@{}}
  [R' |T' ]\\\hline
  [R |T ]
  \end{array}& using \; Eq.\ref{e16}
\end{array}} \right.
\label{e19}
\end{equation}
where (x',y') are the re-projected location of the (x,y) using the correct rotation matrix and translation vector. 
\subsection{Model Reconstruction}\label{mr}
\begin{figure*}[h!t!]
\centering
\centering
{\includegraphics[width=.83\linewidth]{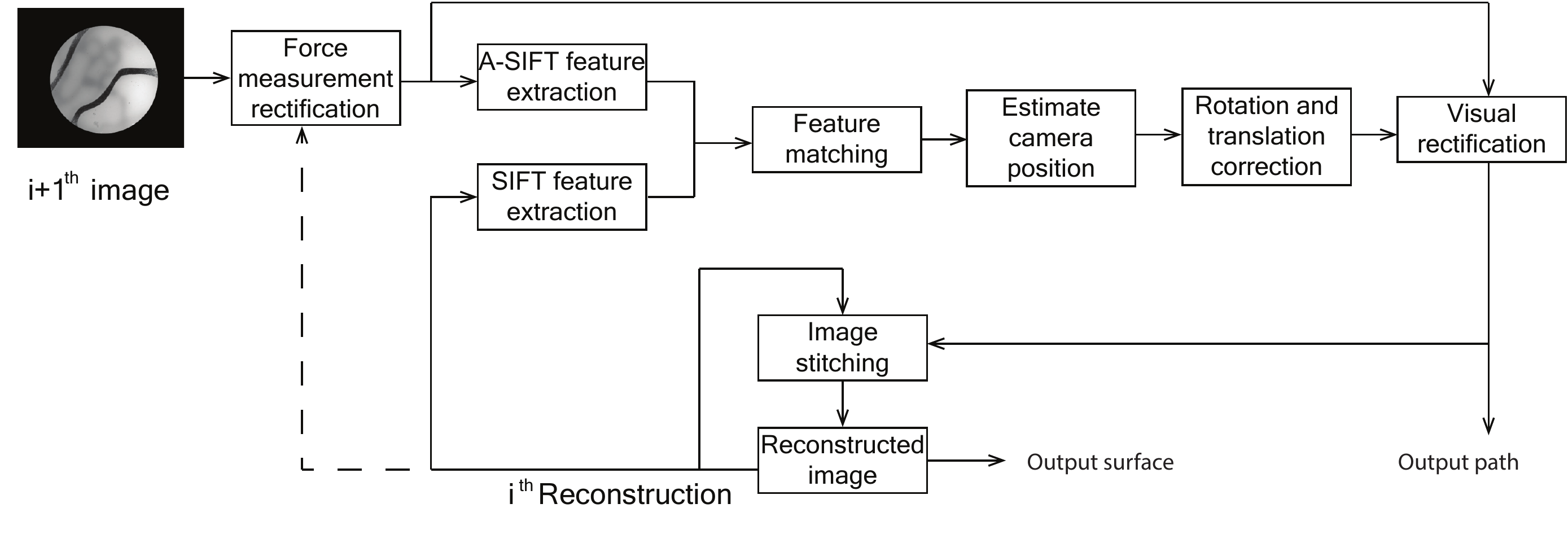}}
\caption{Surface reconstruction and scanning path optimization ($i+1^{th}$ image is compared with $i^{th}$ reconstruction to get the reconstructed surface of $i+1^{th}$ reconstruction.)}
\label{fig10}
\end{figure*}
As discussed earlier, visual based reconstruction alone is not sufficient to rectify the deformation because of the non-steady deformation due to an application of force. In this scenario, only visual reconstruction creates a relative camera position, but stretching of the surface does not make linear relation with the camera position obtained using the visual reconstruction. Which means, the use of visual estimation alone will not be sufficient for non-rigid surface reconstruction.

The force-based reconstruction tries to rectify the structure using the force analysis. But there is $\Delta t$ time required to capture the accurate deformation made by a certain force. Moreover, visual deformation made by the lens could not be reconstructed using the force analysis.
In this scenario, dual force and visual rectification pipelines are necessary to identify the original structure of the surface.

\{$I_i$, $I_2$,..., $I_i$, $I_{i+1}$,...\} is the image set captured using the probe. It is assumed that $I_1$ is the non deformed surface and assigned as the reconstructed sample $RI_1$. In this scenario, $RI_i$ can be defined as the $i^{th}$ reconstruction. The force based deformation model is applied on the $RI_i$ and $I_{i+1}$  to generate rectified $I^\digamma _{i+1}$. 

SIFT and modified A-SIFT are applied on the image $RI_i$ and $I^\digamma _{i+1}$ respectively. SIFT matching, RANSAC and camera pose of the $i+1^{th}$ image is estimated using the reference of $RI_i$. As discussed before, the $I^\digamma _{i+1}$ is the rectified correct camera position to generate the visual and force rectified image $I^{\digamma V} _{i+1}$.
  
Using the (X, Y) translations obtained from the translation matrix, the position of the $I^{\digamma V} _{i+1}$ can be identified with respect to the $RI_i$. Assuming the starting point of $RI_i$, the absolute position of $I^{\digamma V} _{i+1}$ can be estimated. Using the image stitching methods, the next reconstruction instance $RI_i+1$ is generated. The overall visual reconstruction procedure is depicted in Fig.~\ref{fig10}.

\subsection{Accuracy Analysis}
The accuracy of the model depends on the deformation detection and rectification.  Firstly, it has been shown that the calculation of the force using the four pressure sensor provides the total force, the angle of the probe, and the depth of the probe.
Using the force based orientation estimation method, as discussed in Sec.~\ref{fmrm}, the deformation of the view surface, as discussed in Sec.~\ref{def_mod}, can be calculated and the original surface structure is estimated. Hence, the scanning surface force angle can be different (vertical or angular) and the surface position may not be always horizontal, though this is not a problem, as angles are made with respect to the surface. 

The visual rectification by estimation of the probe (camera) position provides the non-deform surface structure. Moreover, the optimized new version of modified A-SIFT reduces the computational time required for A-SIFT. Using the SfM over the optimised matching, obtained by using the modified A-SIFT, will provide a large number of inlier matching to estimate the relative camera position. As the scanning probe will touch the surface for scanning, the assumption of the structure of the surface is not possible by analysing the camera position. Moreover, the camera angle is calculated by visual measurement of the stretching. So the camera angle depends on both force and stretching constants (\textit{Young's modulus} (\textit{E}) and Poisson ratio ($\upsilon$)). Hence, the original probe angle will not be equal with the calculated angle of the visual estimated camera angle, as claimed in Sec.~~\ref{vrm},~\ref{mr}.

In this application, a combination of visual and force model will correctly estimate the surface structure and the probe position (rotation and translation), as the force model provides a unified plane allowing for more matching, and the visual model estimates the structure. In other words, the generated error in the visual reconstruction can be removed using the force reconstruction and the error in the force based reconstruction can be removed using the visual reconstruction. Moreover, the image reconstruction is carried out at the position of the first image (assuming no force is applied). So it can be concluded that the reconstruction will provide the original surface structure accurately.

    

\section{Experiment Set-up}\label{s4}
In this work, the breast phantom and the camera is prepared for scanning purpose. All the required equipment details and calibration methods are discussed in the following subsections.
 
\subsection{Equipment Details}
Here, the model probe is constructed using a visual layer of IR support and four sensors (as shown in Fig.~\ref{fig4}). The 850 nm IR illumination ring is placed in between the camera and the transparent visual layer to capture the vein pattern inside of the skin. 

\begin{figure}[h!t!]
\centering
\centering
\includegraphics[width=.85\linewidth]{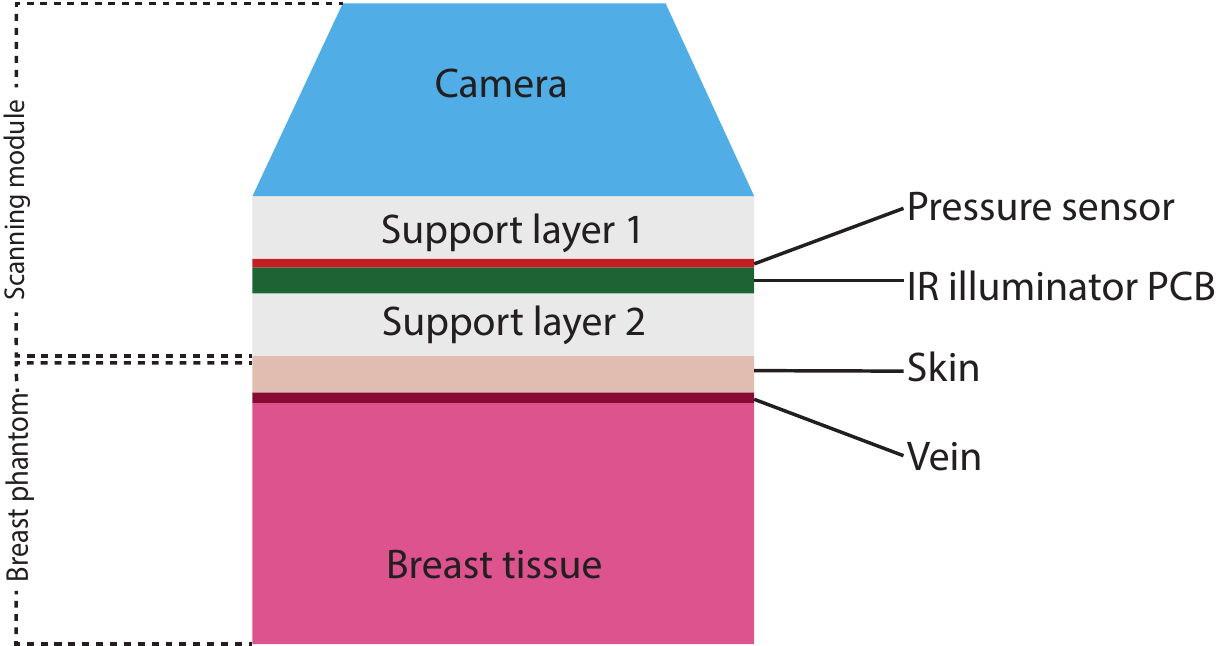}

\caption{Breast model material layers and scanning probe assembly}
\label{fig16}
\end{figure}

For the breast model, Ecoflex 00-10 shore hardness silicone (Smooth-On, US-PA) with 20\% thinner is used to simulate the elasticity of a typical breast~\cite{mechanical,bcancer}. The skin is made-up with the very soft 000-35 shore hardness silicone to simulate the skin stretching. The vein pattern is created using a mixture of silicone and IR absorb material (graphite powder) and placed in between the breast tissue and skin. Fig.~\ref{fig16} shows the layers of the breast model.

A continuous scan is carried out across the surface of the breast phantom and the scanning path is recorded using a  VICON camera tracking system. Here the VICON is constructed using twelve cameras and can measure a global position and orientation with six degree freedom with $\approx$0.5mm position accuracy ~\cite{vicon}. The material of the phantom is very soft (like the breast), so very little pressure will change the surface and stretch the vein pattern like in the real scenario.

\subsection{Probe and Model Calibration}\label{pmc}

\begin{figure}[h!t!]
\centering
\centering
\includegraphics[width=.9\linewidth, height=7cm]{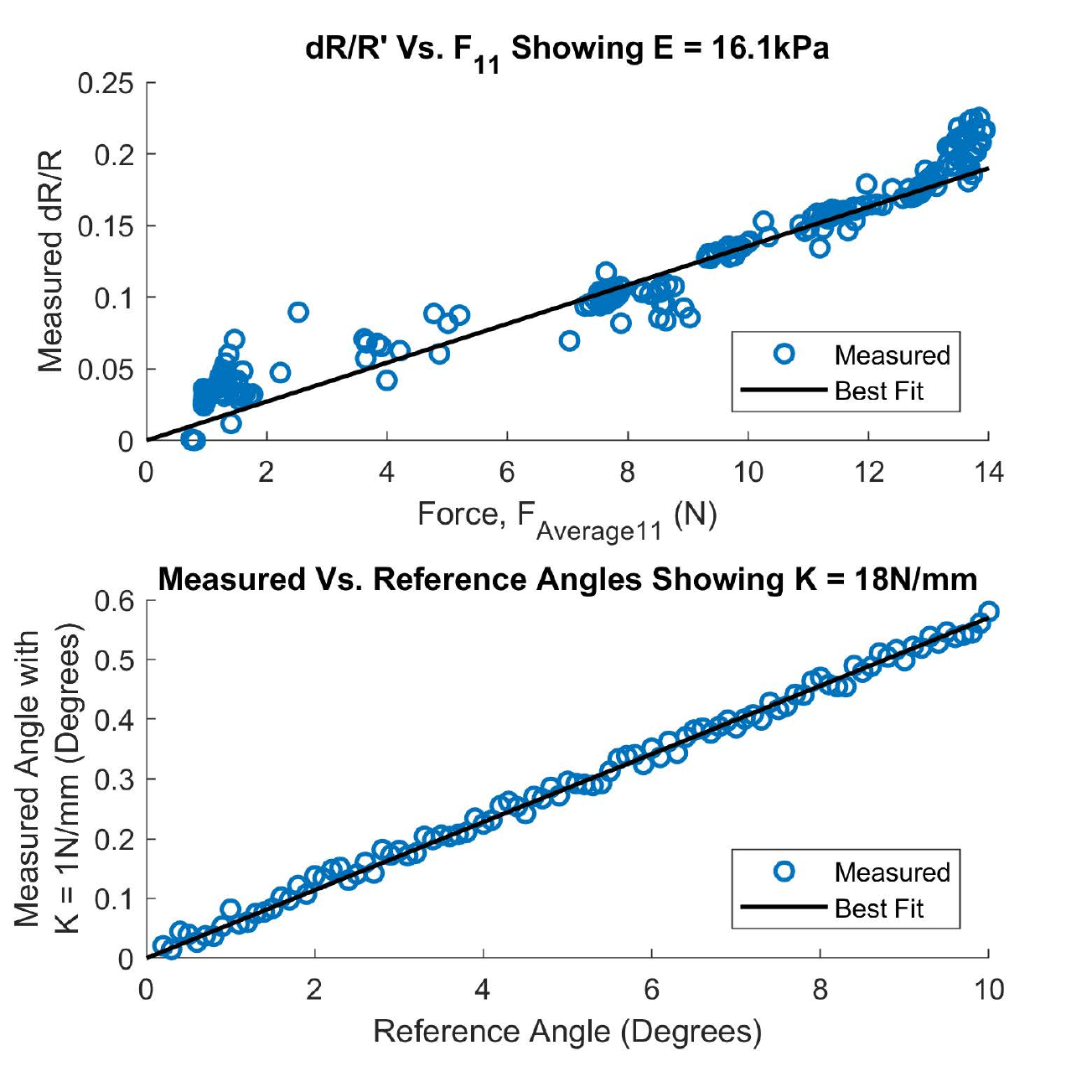}
\caption{\textit{Young’s modulus} and \textit{Hooke's constant} calibration data. E is calculated using measured image feature deviation. K is calculated using a reference angle measurement}
\label{fig12}
\end{figure}

The four pressure sensors that are used to measure the tilt angle using (\ref{e11b}) can also calculate \textit{Hooke's constant}, $\kappa$, with a known reference angle. An Invensense MPU6050 is used to provide a reference measurement for the scanner tilt angle and calibrated using an assumed  $\kappa_{initial}=1N/mm$ for the sensors to identify the real \textit{Hooke's constant}, $\kappa$. Fig.~\ref{fig12} 
shows the calculation of the $\kappa$ using the data points.  Here the measured Hookean constant is estimated as: $\kappa=18N/mm$.
With the scanner area, $\mathcal{A}$, taken to be 0.0048m2 and the material Poisson ratio, $\upsilon$, taken as 0.5 (incompressible), the value of \textit{Young's modulus} (\textit{E}) is $E=16.1KPa$ as shown Fig.~\ref{fig12}. Calibrating the system for different patients is achieved by applying the scanner normally to the tissue, and comparing loaded feature positions with those of an unloaded surface image, as we have done here.
The summary of experimental parameters for the model validation is presented in the Table~\ref{tset}.

\begin{table}[!ht]
	\caption{Experiment set-up.\label{tset}}
	\centering{
	\centering
\resizebox{.9\linewidth}{!}
{		
\def\arraystretch{1.15}
		\begin{tabular}{ |c|c|c|c|c|c| }
		
\hline
\rotatebox{0}{Image}&\rotatebox{0}{Sensor}&Poisson&\rotatebox{0}{Hooke's}&\rotatebox{0}{Scanner }&Young's\\
set& distance&ratio$\upsilon$&constant&area&modulus\\\hline
336&53mm&0.05&18N/mm&0.0048$m^2$&16.1 KPa\\

\hline

\end{tabular}}
}	
\end{table}


Here the stretching constant can be calculated in the beginning of the clinical testing. When the probe touches the surface the first set of images will be used for calculating the stretching constant and to select the first image of the experiment. 

\section{Results}\label{s5}
\begin{figure}[h!t!]
\centering
\centering
\includegraphics[width=.8\linewidth]{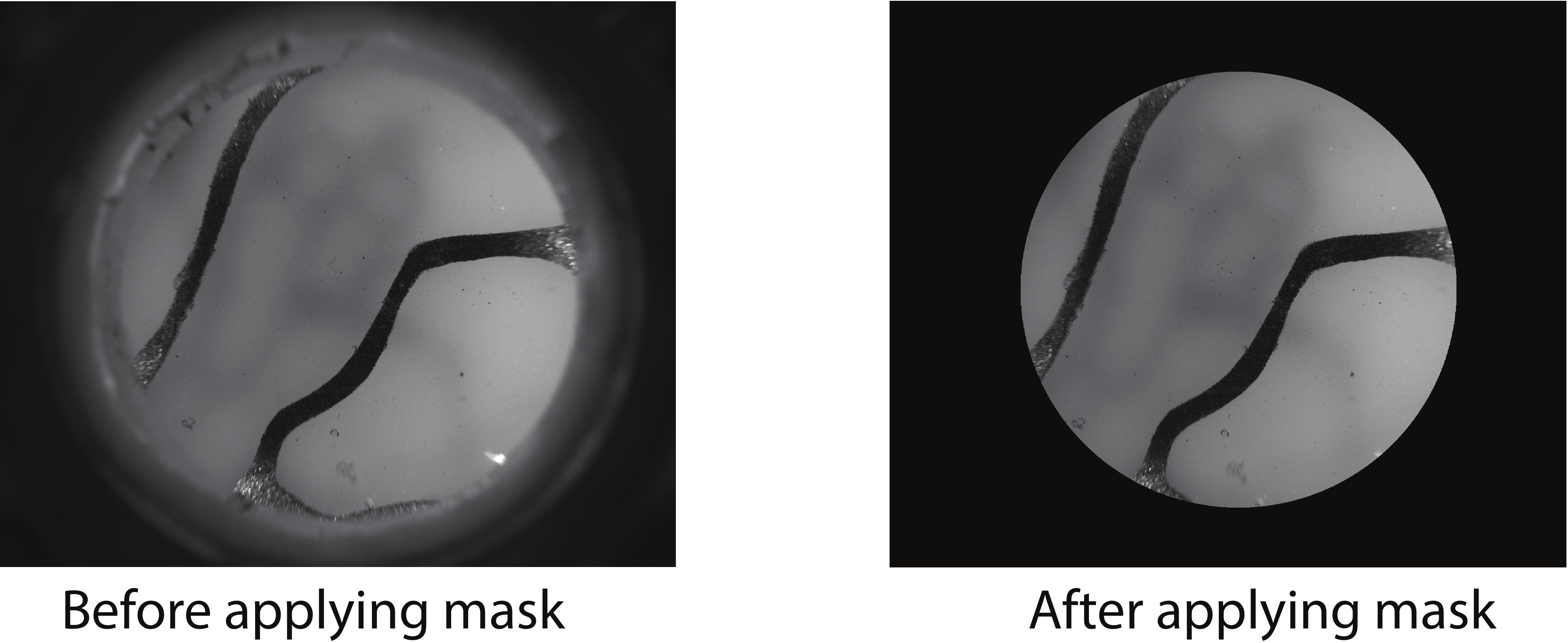}

\caption{Rectification and vein part selection using mask}
\label{fig20}
\end{figure}

In this experiment, the images are pre-filtered by selecting the scanner area using a mask to prepare for the experiment as shown in Fig~\ref{fig20}.
 
\begin{figure}[h!t!]
\centering
\includegraphics[width=.9\linewidth]{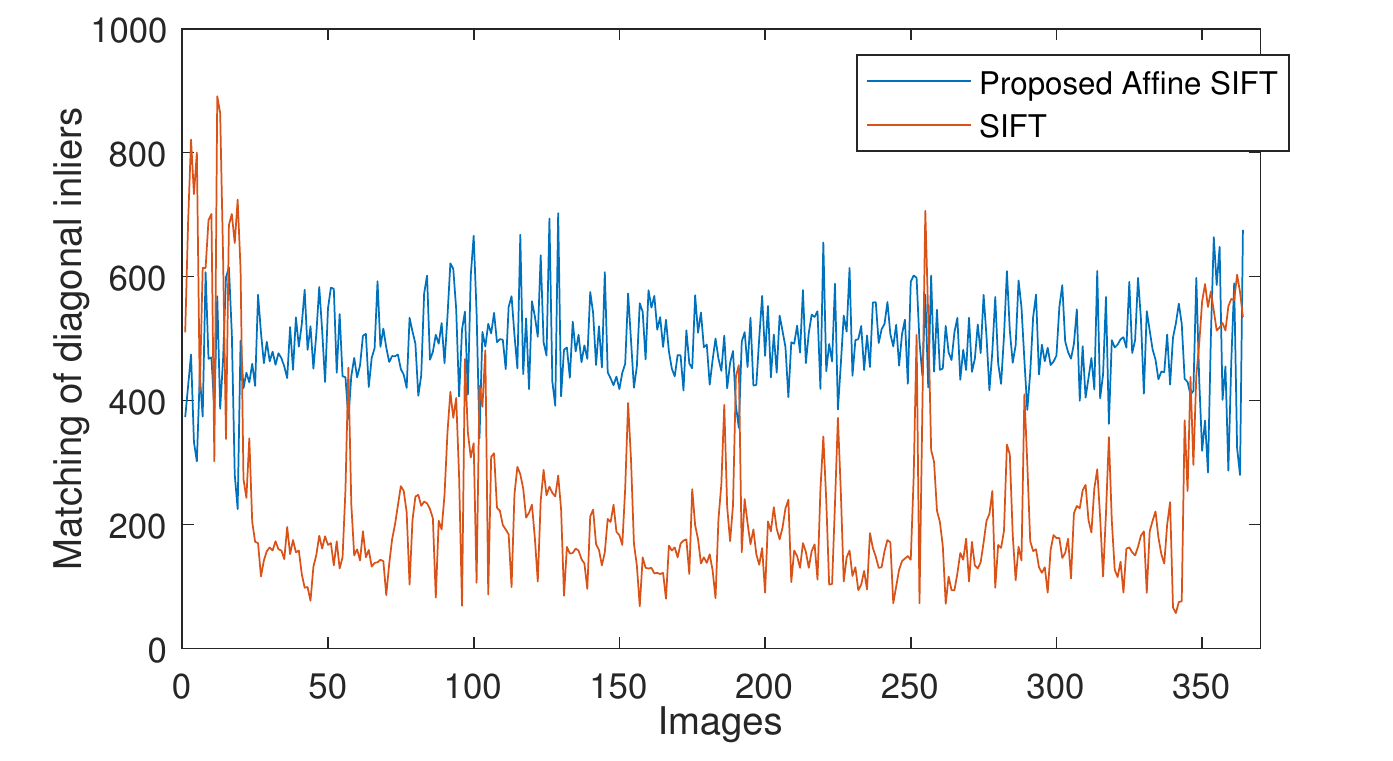}


\caption{Matching efficiency comparison between modified A-SIFT and SIFT for sequential inlier matching }
\label{fig141}
\end{figure}

 After that, the modified A-SIFT feature estimation is carried out to find out the inlier matching. For better matching SIFT parameters are set such a way that only the vein pattern regions are selected for feature extraction. Peak threshold 1, edge threshold 10 is used for modified A-SIFT matching. Fig.~\ref{fig141} shows that the use of  modified A-SIFT increases the inlier matching  for the breast images in sequential matching by well over 100\%.
\subsection{Model Construction Result}

\begin{figure}[h!t!]
\centering
\centering

\subfigure[Inliers matching of image sequences]{\includegraphics[width=.43\textwidth]{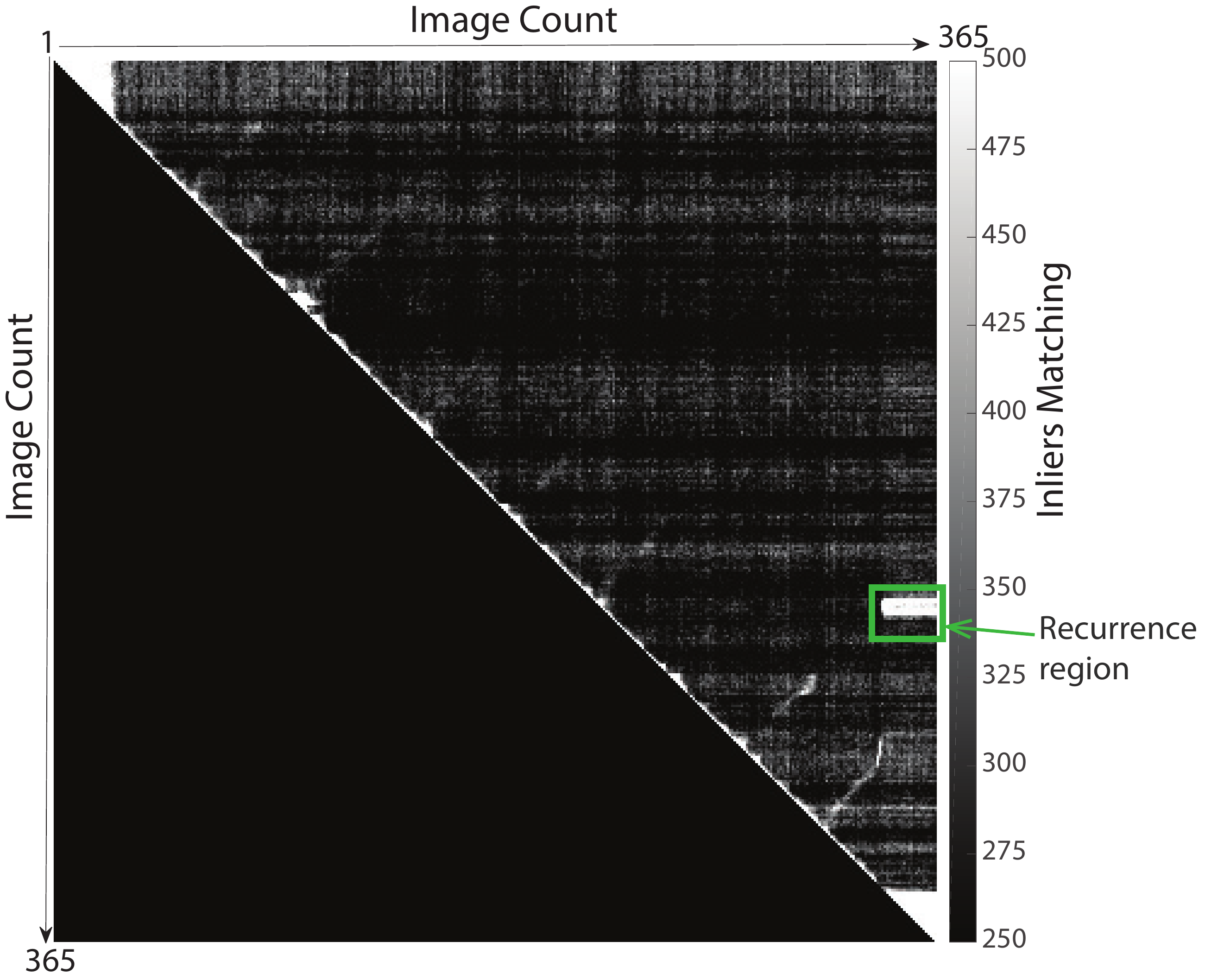}\label{fig13}  }
\subfigure[Comparison of extracted camera movement path with real VICON data]{\includegraphics[width=.45\textwidth]{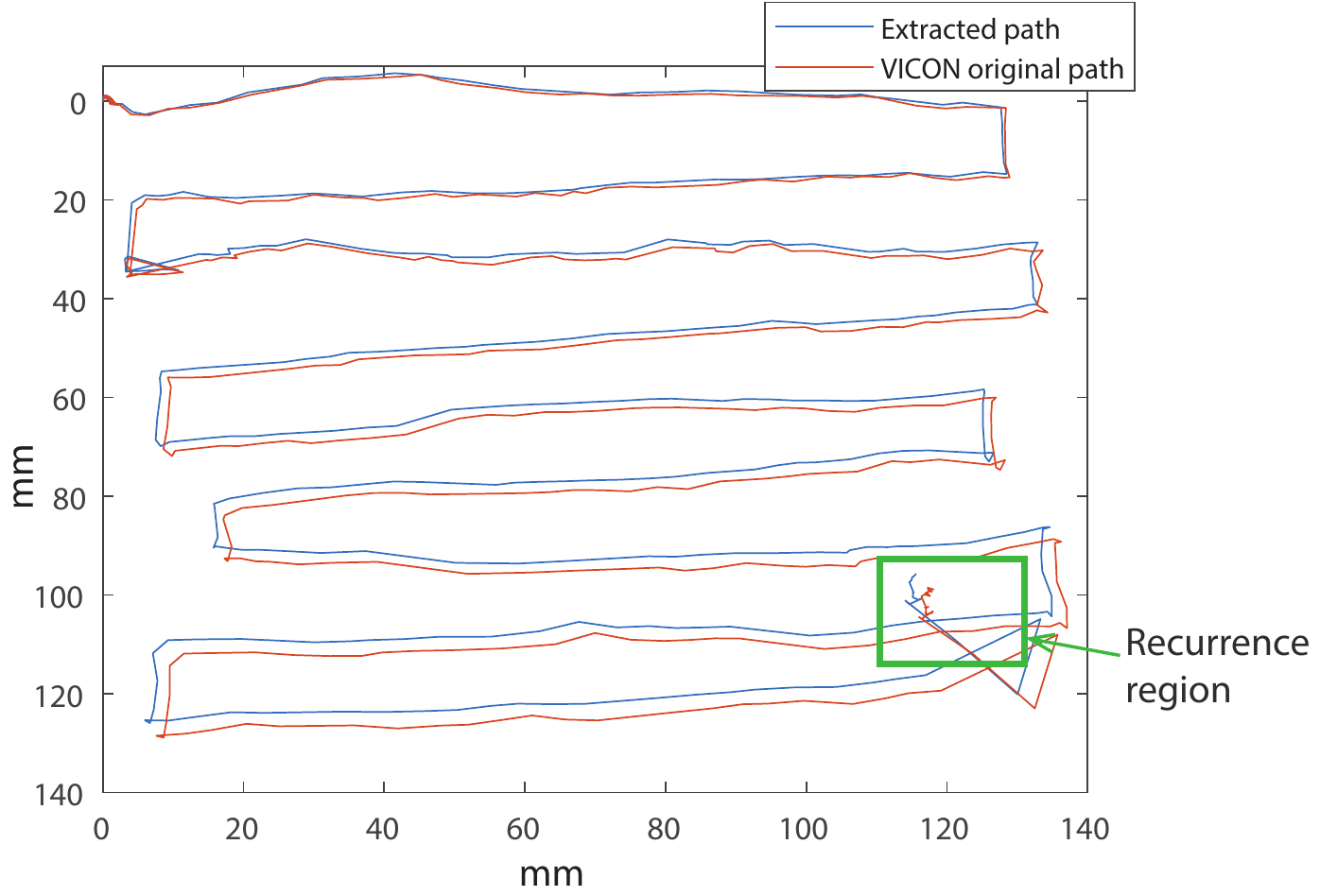}\label{fig14} }
\caption{Inliers matching and scanning path estimation of the proposed scheme, where the green marked regions shows the recurrence region during scanning}
\label{fig1401}
\end{figure}

In this work, the scanning is carried out in a sequential order.
 Hence, the majority of correct A-SIFT matching will occur in the next sequence of the image sets. Fig.~\ref{fig13}
 shows the inlier matching with sequential images. It is observed that some better matching appears in intermediate scanning because the random movement of the probe. From the image, it can be concluded that the use of the Modified A-SIFT improves the matching efficiency for the image set over conventional SIFT.

The translation matrix is generated using the sequential inlier matching as shown in Fig.~\ref{fig13}.
In the proposed method, the (X,Y) components of the translation matrix defines the position of the camera. The resultant position in the experiment is compared with the real position obtained using the VICON camera tracking system. The matching result is shown in Fig.~\ref{fig14}.

The camera position detection of the proposed scheme is compared with the latest non-rigid reconstruction methods~\cite{sfm,agudo3,nrsfm1}. 
The authors made use of the detail explanations from research papers Ref.~\cite{agudo3} and ~\cite{nrsfm1} (as explained in the literature) for the separate implementations. Here the approach of this research work is different than the existing works. So, the validations of the implementations are carried out by using the similar datasets (as in~\cite{sfm,agudo3,nrsfm1}) and similar results have been achieved to those claimed by Ref.~\cite{sfm,agudo3,nrsfm1}.
Also, the claim in Fig~\ref{fig13} for the better matching at the sudden stage is clearly visible in the scanning path as indicated by the green marked region. The result of the scheme for 1274 mm camera travel path is tabulated in the Table~\ref{tpos}. 
\begin{table}[!ht]
	
	\caption{Camera position estimation comparison of the proposed scheme with existing schemes.\label{tpos}}
	\centering{
	\centering
\resizebox{\linewidth}{!}
{		
\def\arraystretch{1.15}
		\begin{tabular}{ c|c|c|c|c| }
		
\cline{2-5}
& Proposed&SfM\cite{sfm}&FEM\cite{agudo3}&NRSFM\cite{nrsfm1}\\\hline
\multicolumn{1}{|c|}{RMSE Error (mm)}&2.261&15.263&11.821&12.36\\ \hline
\multicolumn{1}{|c|}{Error ratio (\%)}&0.296&1.833&1.496& 1.221\\
\hline

\end{tabular}}
}	
\end{table}

\begin{figure}[h!t!]
\centering
\includegraphics[width=.9\linewidth]{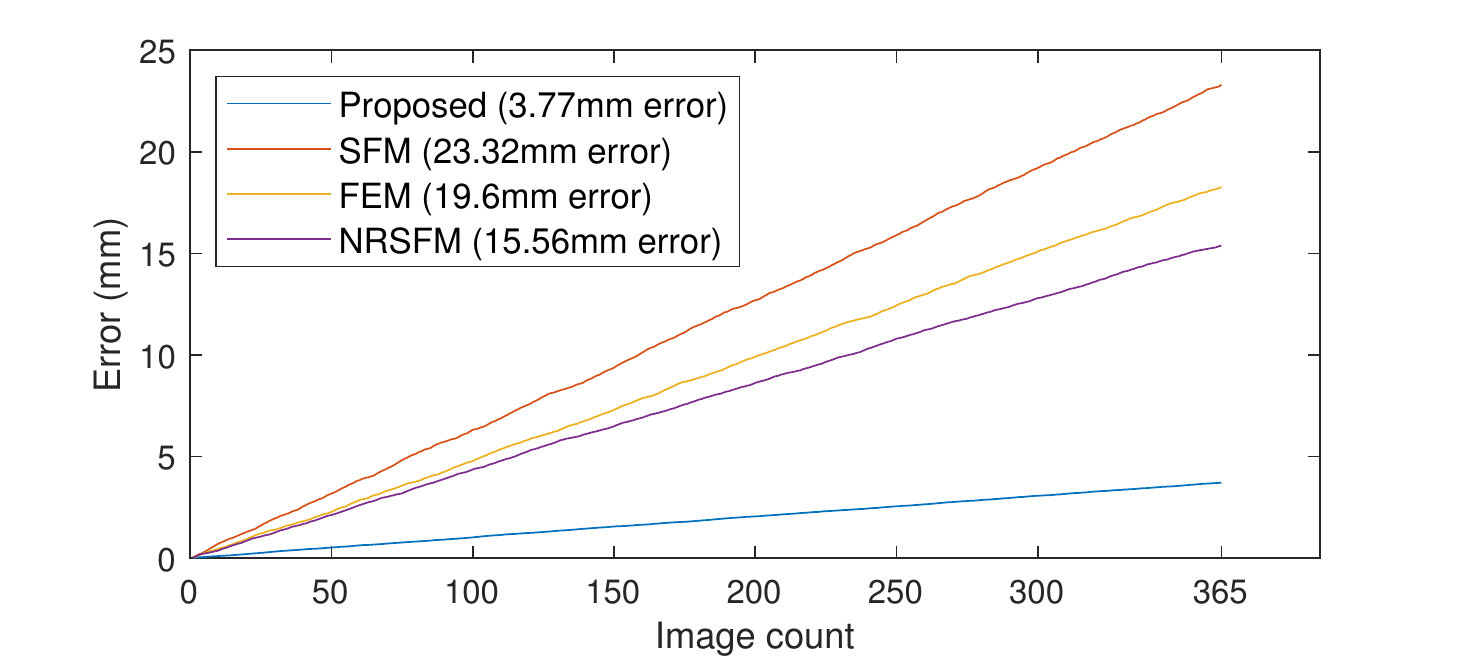}


\caption{Camera position estimation error comparison with the increasing of images for proposed scheme and existing schemes \cite{sfm,agudo3,nrsfm1}}
\label{fig161}
\end{figure}

Here the average error is the root mean square error (RMSE) in mm.  The error ratio defines the total propagated error ratio as shown in (\ref{e22}).
\begin{equation}
Error\;ratio=100\left(\frac{Total\;error}{Total\;distance\;travelled}\right)
\label{e22}
\end{equation}
The total error in time of scanning with respect to the increment of image sequence is shown in Fig.~\ref{fig161}. The error is calculated by measuring the linear distance from the VICON~\cite{vicon} position and estimated position in each schemes.

\begin{figure}[h!t!]
\centering

\subfigure[Steady force is applied (increasing and decreasing of force)]{\includegraphics[width=.9\linewidth]{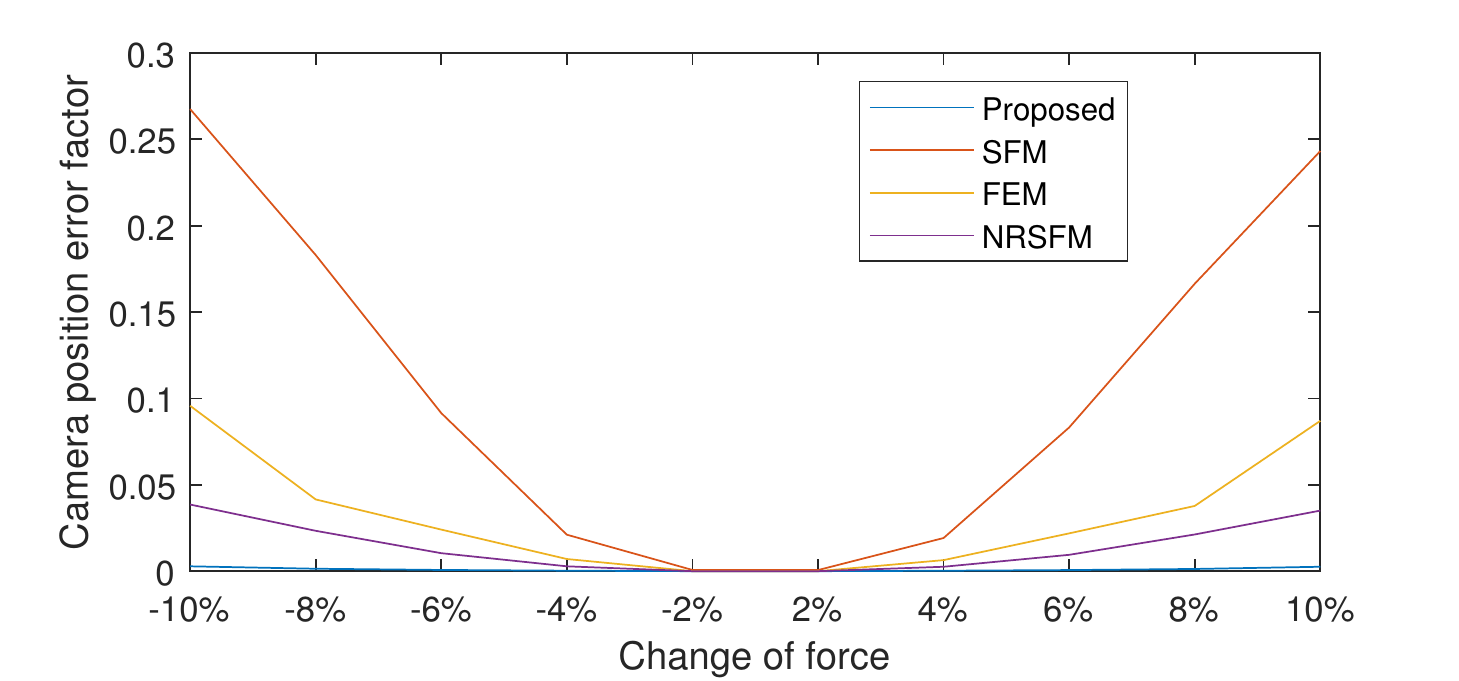}  \label{aaaa1}}
\subfigure[Angular affine force applied (with a ratio of $\frac{F1-F2}{F1}$ along the X axis)]{\includegraphics[width=.9\linewidth]{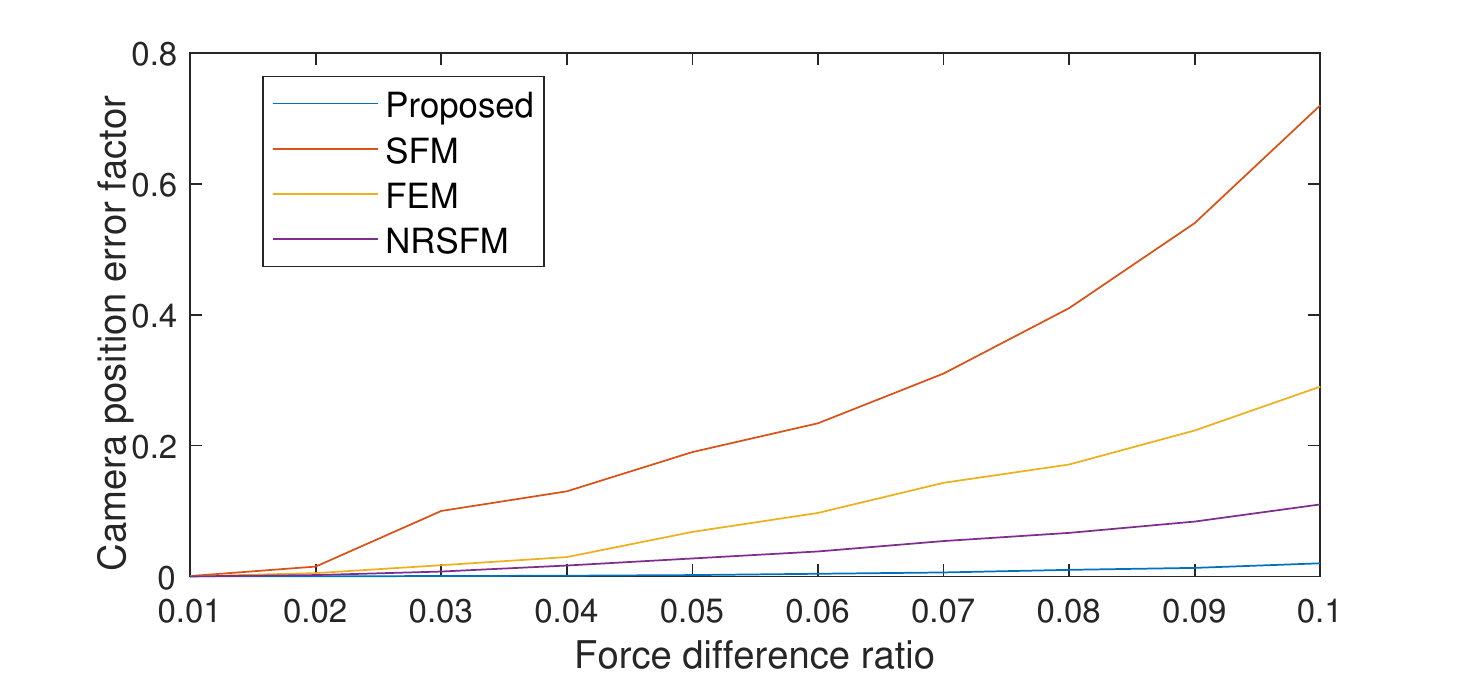}  \label{aaaa2}}

\caption{Camera position error factor with the application of force}
\label{fig171}
\end{figure}

\begin{figure}[h!t!]
\centering
\centering

\subfigure[Original]{\includegraphics[width=.330\textwidth]{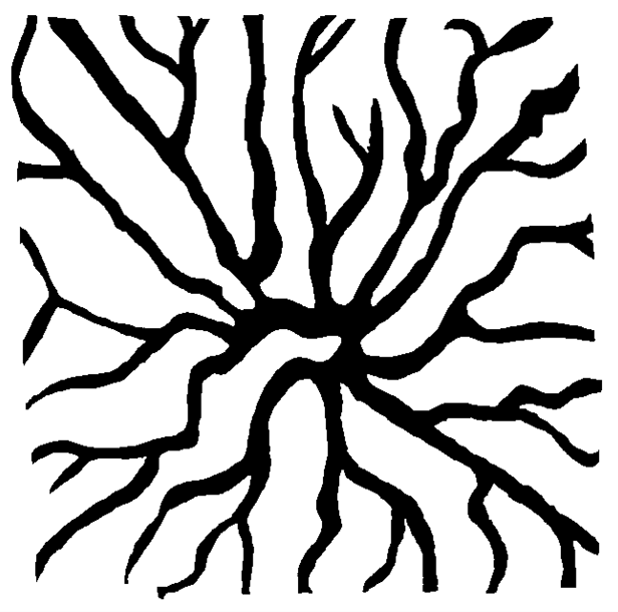}  \label{15_1}}
\subfigure[Reconstructed]{\includegraphics[width=.330\textwidth]{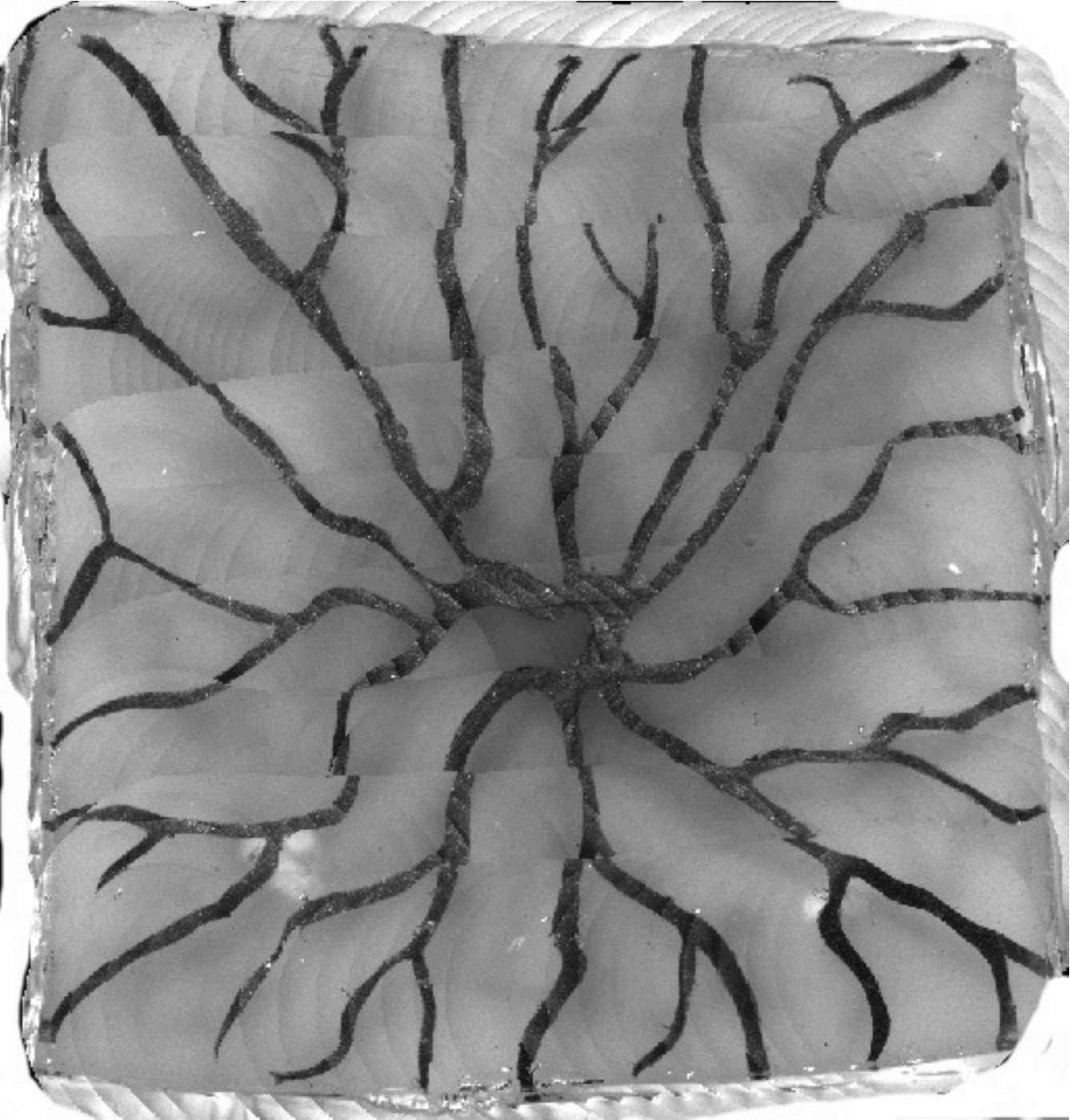}  \label{15_2}}
\caption{Visual comparison of original and reconstructed surface }
\label{fig15}
\end{figure}

The error in the camera position is occur due to the stretching deformation. A clear view of camera position estimation is shown in Fig.~\ref{fig171}. Here the camera position estimation error factor is calculated by the relative camera position translation vector compared original camera position vector obtained from VICON~\cite{vicon}. The camera position error factor is calculated as $(\frac{error\; displacement\; in\; estimation}{original\; displacement\; (VICON)})$. 
 It is clear that the camera position estimation is close to accurate for the proposed scheme than the existing schemes~\cite{sfm,agudo3,nrsfm1}. The better camera position provides a better reconstruction surface of the breast phantom.
Using the scanning path as shown in Fig.~\ref{fig14}, the reconstruction is carried out.
  Fig.~\ref{15_2} shows the reconstructed surface of the vein pattern for the original reference of Fig.~\ref{15_1}.

It is clearly observed that the proposed scheme creates an identical vein pattern of the reference. Moreover, the scaling and stretching effect does not adversely effect this reconstruction due to the efficient visual and force based model analysis.

 It can be claimed from the experiment that the proposed scheme gives an efficient non-rigid reconstruction method for model construction and image registration. 
\subsection{Discussion}
In this work, the images are rectified using the force and visual reconstruction methods to make a surface model of a breast phantom using the non-rigid images. The reconstruction procedure has three major contributions to knowledge for efficient reconstruction. 
Firstly using the force analysis and angular measurement, reduce the contact based deformation. The calculation of accurate \textit{Young's modulus} and \textit{Hooke's constant} gives the accurate stretching factor as explained in Eq.~\ref{e15}. 
Secondly, the use of Visual reconstruction methods by analysing the camera position obtained using SfM allows for correction of camera position and re-projection using the rectified position, improving the accuracy of the reconstruction.
Finally, modified A-SIFT improves the matching efficiency of the proposed scheme compared with SIFT. Also, the accurate angle measurement and proper selection of the affine coefficients, make the coefficients much faster than the traditional method. The traditional A-SIFT feature extraction works as $\approx13\times$ and matching takes $\approx169\times$. In this modified A-SIFT the feature extraction and matching takes $\approx9.46\times$. Moreover using the A-SIFT improves the accuracy of the SfM approach to optimize the camera position as well as better reconstruction where SfM alone makes an incomprehensible structure.

The stretching constraints of the experiment model is calculated at the  when the scanning probe touch the surface. As a result the variation of Breast elasticity for different patient does not make any restriction for getting accurate localization.
Also, these advances in image mosaicing, when applied to tactile imaging of breast lesions will allow for comprehensive global pressure maps to be produced, which will allow for greater diagnostic potential in the future.

\section{Conclusions}\label{s6}
In this paper, a medical image construction and registration technique has been proposed for segmented images scanned using a prototype model of a breast scanning probe. The proposed force based rectification model, removes the stretching deformation due to the contact force. The newly modified affine SIFT feature estimation technique finds the spatial feature points from the deformed surface for model construction. The combination of the both reconstruction models provide a nearly flawless solution for camera position estimation and image reconstruction technique.
Additionally this type of system will work for any body tissues with comprehensive vascular networks and can be extended to clinical examination of tissues such as the abdomen.

\section*{Acknowledgment}

The authors would like to thank EPSRC for funding ``Improvement of Breast Cancer Tactile Imaging through Non-Rigid Mosaicing" project through grant number EP/P011276/1. Also would like to thank Sure Inc. and PPS Inc. for supporting this project with $SingleTact$\textsuperscript{TM} pressure sensors.


%

\bibliographystyle{IEEEtran}
\bibliography{reference}

\end{document}